\pdfoutput=1
\documentclass[11pt]{article}

\usepackage[final]{acl}

\usepackage{times}
\usepackage{latexsym}
\usepackage{amsmath}
\usepackage{enumitem}

\usepackage[T1]{fontenc}
\usepackage[utf8]{inputenc}

\usepackage{microtype}
\usepackage{subcaption}
\usepackage{inconsolata}
\usepackage{multirow}
\usepackage{float}
\usepackage{graphicx}
\usepackage[capitalize]{cleveref}
\crefname{section}{Sec.}{Secs.}
\Crefname{section}{Section}{Sections}
\Crefname{table}{Table}{Tables}
\crefname{table}{Tab.}{Tabs.}
\Crefname{figure}{Figure}{Figures}
\crefname{figure}{Fig.}{Figs.}
\title{Redundancy Principles for MLLMs Benchmarks}

\author{
 \textbf{Zicheng Zhang\textsuperscript{1,2}\footnotemark[1]},
 \textbf{Xiangyu Zhao\textsuperscript{1,2}\footnotemark[1]},
 \textbf{Xinyu Fang\textsuperscript{1,3}},
 \textbf{Chunyi Li\textsuperscript{1,2}},
 \textbf{Xiaohong Liu\textsuperscript{2}},
\\
 \textbf{Xiongkuo Min\textsuperscript{2}},
 \textbf{Haodong Duan\textsuperscript{1}\footnotemark[2]},
 \textbf{Kai Chen \textsuperscript{1}\footnotemark[2]},
 \textbf{Guangtao Zhai\textsuperscript{1,2}\footnotemark[2]},
\\
 \textsuperscript{1}Shanghai AI Laboratory,
 \textsuperscript{2}Shanghai Jiaotong University,
 \textsuperscript{3}Zhejiang University
}

\begin{document}
% \maketitle
\twocolumn[{%
\renewcommand\twocolumn[1][]{#1}%
\maketitle
\vspace{-15mm}
\begin{center}
    \centering
    \captionsetup{type=figure}
    \includegraphics[width=0.58\linewidth]{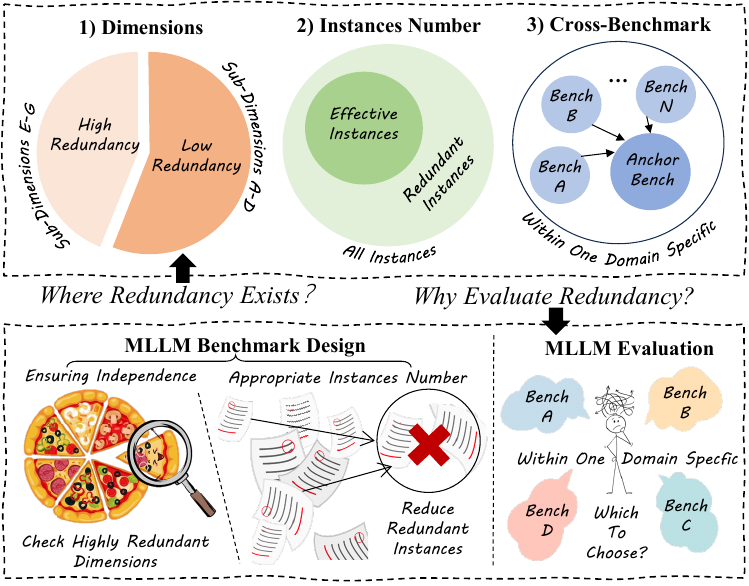}
    \vspace{-2mm}
    \caption{\textit{Where Redundancy Exists?} and \textit{Why Evaluate Redundancy?}}
    %\vspace{2pt}
    \label{fig:redundancy}
\end{center}%
}
]

\begin{abstract}
With the rapid iteration of Multi-modality Large Language Models (MLLMs) and the evolving demands of the field, 
the number of benchmarks produced annually has surged into the hundreds. 
The rapid growth has inevitably led to significant redundancy among benchmarks. 
Therefore, it is crucial to take a step back and critically assess the current state of redundancy and propose targeted principles for constructing effective MLLM benchmarks.
In this paper, we focus on redundancy from three key perspectives:
\textbf{1)} Redundancy of benchmark capability dimensions, 
\textbf{2)} Redundancy in the number of test questions, 
and \textbf{3)} Cross-benchmark redundancy within specific domains.
Through the comprehensive analysis over hundreds of MLLMs' performance across more than 20 benchmarks, we aim to quantitatively measure the level of redundancy lies in existing MLLM evaluations, provide valuable insights to guide the future development of MLLM benchmarks,
and offer strategies to refine and address redundancy issues effectively.
\end{abstract}

% \renewcommand{\thefootnote}{}
% \footnotetext{This project is supported by the National Key R\&D Program of China (No.2022ZD0161600), the Shanghai Postdoctoral Excellence Program (No.2023023), China Postdoctoral Science Fund (No.2024M751559), and Shanghai Artificial Intelligence Laboratory.}

\footnotetext[1]{Equal Contribution.}
\footnotetext[2]{Corresponding Author.}

\section{Introduction}
\label{sec:intro}

%% needs a brief intro
% MLLM Benchmark Related Works.
% In recent years, with the rapid advancement of MLLMs, there has been an explosive growth in Visual Question Answering (VQA) Benchmarks. 

Model Evaluation has always played a crucial role in the development of Multi-modal Large Language Models (MLLMs). 
Benchmarks serve not only as tools for assessing model accuracy but also as catalysts for driving innovation and improvements within the field. 
In the early stages, traditional model evaluation benchmarks such as GQA~\cite{hudson2019gqa}, VQA-V2~\cite{antol2015vqa}, VizWiz~\cite{bigham2010vizwiz}, and TextVQA~\cite{singh2019towards} are characterized by relatively simple questions and answers, with responses often being a single word. This limits the depth of understanding and reasoning required from the models, making them less effective at evaluating the complex capabilities of modern MLLMs that are expected to handle more nuanced tasks. 
With the emergence of more powerful MLLMs~\cite{achiam2023gpt, team2023gemini, chen2024internvl, wang2024qwen2, li2024llava, liu2024visual}, traditional evaluation frameworks have become inadequate to meet the flexible evaluation requirements. 
In response, a new generation of VQA benchmarks has arisen, such as MMBench~\cite{liu2025mmbench}, MMVet~\cite{yu2023mm}, and MMMU~\cite{yue2024mmmu}.

As MLLMs have rapidly iterated and evolved, their diverse capabilities across various domains have garnered increasing attention, which has led to the development of specialized benchmarks to evaluate MLLMs' performance in specific areas, like 
% \textit{General Understanding} MME~\cite{fu2024mmecomprehensiveevaluationbenchmark}, SEEDBench~\cite{li2024seed}, BLINK~\cite{fu2025blink}, and MMStar~\cite{chen2024we},
% \textit{Mathematics Task}, benchmarks such as MathVista~\cite{lu2023mathvista}, MathVerse~\cite{zhang2025mathverse}, MathVision~\cite{wang2024measuring}, and Dynamath~\cite{zou2024dynamath},
% \textit{Optical Character Recognition (OCR)}, benchmarks like OCR-VQA~\cite{mishra2019ocr}, OCRBench~\cite{liu2023hidden}, and DOC-VQA~\cite{mathew2021docvqa},
% General Understanding ~\cite{fu2024mmecomprehensiveevaluationbenchmark,li2024seed,fu2025blink,chen2024we},
Mathematics Task~\cite{lu2023mathvista,zhang2025mathverse,wang2024measuring,zou2024dynamath},
Optical Character Recognition (OCR)~\cite{mishra2019ocr,liu2023hidden,mathew2021docvqa},
Medical Field~\cite{hu2024omnimedvqa}, Remote Sensing~\cite{li2024vrsbench}, Agents~\cite{yang2024gpt4tools}, GUIs~\cite{baechler2024screenai}, and so on.

The rapid proliferation of benchmarks has inevitably introduced significant redundancies, 
with overlapping capabilities being assessed and recurring questions appearing within and across benchmarks. 
Such redundancies create inefficiencies in model evaluation, repeatedly testing similar aspects of MLLM performance without contributing meaningful new insights. 
Additionally, this trend risks overemphasizing certain task types while neglecting others, 
potentially distorting research priorities. 
In this work, we address these challenges through a comprehensive and systematic exploration.

\subsection{Identifying Redundancy }
Redundancy is an intrinsic and multifaceted issue in benchmarks, appearing in several key forms:

\begin{itemize}[left=0pt, itemsep=0pt, topsep=0pt, parsep=0pt]
    \item \textbf{Redundancy across dimensions (intra-bench):} Tasks within the same benchmark may evaluate overlapping capabilities of MLLMs, leading to repetitive assessments.
    \item \textbf{Redundancy among instances (intra-bench):} Certain instances closely resemble others, providing minimal additional differentiation or insight for model evaluation.
    \item \textbf{Redundancy across benchmarks within specific domains:} Benchmarks targeting specific domains often exhibit overlapping objectives or scopes, 
    resulting in duplicated efforts across different evaluation sets.
\end{itemize}

\subsection{Ideal Redundancy Principles}

Effective benchmarks should adhere to the following principles regarding redundancy:

\begin{itemize}[left=0pt, itemsep=0pt, topsep=0pt, parsep=0pt]
    \item \textbf{Independence of dimensions:}  
    Ideal benchmarks should ensure that its dimensions are largely independent, 
    minimizing overlap between them. 
    However, some degree of redundancy may be inevitable when certain capabilities naturally require the interaction of multiple foundational skills,
    and redundancy should be carefully balanced to avoid excessive overlap.
    
    \item \textbf{Optimal instance count:}  
    A well-designed benchmark should strike a balance in the number of instances: neither too few nor too many, to ensure reliable and meaningful evaluations without introducing unnecessary redundancies.
    
    \item \textbf{Domain representativeness:}  
    A comprehensive benchmark targeted to a specific domain should represent the domain. 
    This may involve purposeful overlap with other benchmarks within the same domain to reflect shared core capabilities.
\end{itemize}

\subsection{Benifits of Evaluating Redundancy}

Evaluating and addressing redundancy offers several significant benefits, as shown in \cref{fig:redundancy}:

\begin{itemize}[left=0pt, itemsep=0pt, topsep=0pt, parsep=0pt]
    \item \textbf{Optimizing benchmark design:} \textbf{1).} Determines whether certain dimensions within a benchmark warrant separate assessments or can be consolidated; \textbf{2).} Identifies the minimal and sufficient number of instances required for accurate evaluation; \textbf{3).} Assesses the necessity of introducing new benchmarks within specific domains.
    
    \item \textbf{Enhancing efficiency in MLLM evaluation:} 
    \textbf{1).} Determines whether a benchmark deviates from the domain's distribution ;
    \textbf{2).}  Identifies the anchor benchmarks required to evaluate model performance within the domain.
\end{itemize}

By systematically addressing redundancy, we can not only enhance the principles of benchmark design but also alleviate the resource demands of MLLM evaluation, 
creating a more streamlined and effective evaluation ecosystem.

\section{Redundancy Framework}
We present a framework for evaluating redundancy among MLLM capabilities, defined as specific tasks within a benchmark.
Our framework is grounded in the following prior assumption: 

\textit{
When evaluating similar capabilities, 
the performance rankings of MLLMs should exhibit strong correlation. 
Conversely, significant differences in these rankings suggest the evaluated capabilities are relatively independent. }

Based on this principle, 
we propose the \textbf{Performance Correlation Redundancy Framework}, 
which quantifies redundancy by measuring the correlation of MLLM performance rankings. 
To ensure robustness and generalization capability, 
we leverage the comprehensive data from VLMEvalKit~\cite{duan2024vlmevalkit}, 
which includes diverse benchmarks and performance results from more than 100 MLLMs.

\subsection{Dimensions Redundancy}

Assume a benchmark consists of a set of dimensions, denoted as \( X = \{X_1, X_2, \ldots, X_m\} \), where each \( X_i \) represents a specific dimension. Let \( N \) denote the number of MLLMs evaluated on these dimensions. For a given dimension \( X_i \), we denote the ranking of the \( N \) MLLMs on this dimension as \( R_i \).
To quantify the redundancy of \( X_i \), we compute the average rank correlation between \( R_i \) and the rankings \( R_j \) of all other dimensions \( X_j \) (\( j \neq i \)). Formally, the redundancy \( \rho(X_i) \) is defined as:
\begin{equation}
    \rho(X_i) = \frac{1}{m-1} \sum_{\substack{j=1 \\ j \neq i}}^m \text{CORR}(R_i, R_j),
\end{equation}
where \( \text{CORR}(R_i, R_j) \) is the correlation coefficient between the rankings \( R_i \) and \( R_j \).
\begin{itemize}[left=0pt, itemsep=0pt, topsep=0pt, parsep=0pt]
    \item High $\text{CORR}(R_i, R_j)$ values can help identify potentially redundant dimension pairs.
    \item \( \rho(X_i) \) represents the average redundancy level of dimension $X_i$, quantifying its overall overlap. 
\end{itemize}

By calculating the redundancy \( \rho(X_i) \) for all dimensions \( X_i \) in the benchmark and averaging these values, we can obtain the overall internal redundancy of the benchmark as well. Formally, the benchmark internal redundancy \( \rho_{BI} \) is defined as:
\begin{equation}
\rho_{BI} = \frac{1}{m} \sum_{i=1}^m \rho(X_i),
\end{equation}
where \( \rho(X_i) \) is the redundancy of the \( i \)-th dimension as previously defined.
This metric reflects the average similarity among all dimensions within the benchmark. A lower \( \rho_{BI} \) suggests that the dimensions are relatively independent and diverse.

\begin{figure}[tbp]
    \centering
    \includegraphics[width=0.85\linewidth]{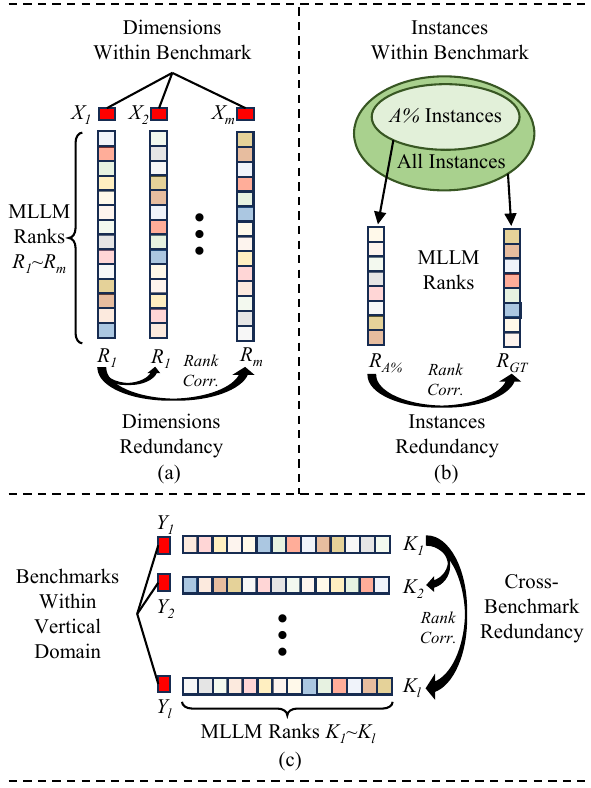}
    \caption{A quick look at the redundancy framework, where (a), (b), and (c) show the general process of computing dimensions redundancy, instances redundancy, and cross-benchmark redundancy respectively.}
    \label{fig:framework}
    \vspace{-10pt}
\end{figure}

\subsection{Instances Redundancy}

Let a benchmark contain \( M \) total instances (\textit{e.g.}, QA pairs). 
To evaluate redundancy, we begin by calculating the MLLM performance rankings obtained over the full set of all \( M \) instances, 
denoted as the ground-truth ranking \( R_{\text{GT}} \).
We then randomly sample a subset of the instances, comprising A\% of the total M, 
and compute the corresponding MLLM rankings, 
denoted as \( R_{\text{sample}} \). 
To quantify the redundancy of the benchmark at a sampling ratio of \( A\% \), 
we calculate the correlation coefficient between \( R_{\text{sample}} \) and \( R_{\text{GT}} \).
This correlation reflects how representative the sampled subset is of the entire benchmark. 
To reduce the effect of randomness, the sampling process is repeated \(T = 100 \) times, 
and the average correlation result is recorded.
We define the instance redundancy of the benchmark at sampling ratio \( A\% \), 
denoted as \( \rho(A\%) \), as follows:
\begin{equation}
    \rho(A\%) = \frac{1}{T} \sum_{1 \le t \le T} \text{CORR}(R_{A\%_{t}}, R_{\text{GT}}),
\end{equation}
where \( R_{A\%_{t}} \) represents the MLLM ranking based on the sampled \( A\% \) instances at the \( t_{th}\) time, 
and \( R_{\text{GT}} \) is the MLLM ranking based on the full \( M \) instances within the MLLM benchmark. 
The interpretation of \( \rho(A\%) \) is straightforward: 

\begin{itemize}[left=0pt, itemsep=0pt, topsep=0pt, parsep=0pt]
\item A higher \( \rho(A\%) \) indicates that the sampled instances are highly representative of the entire benchmark, 
and the remaining \( 1-A\%\) instances contribute little additional information.
\item Conversely, a lower \( \rho(A\%) \) suggests that the sampled instances are less representative, 
and more instances are needed to capture the variability of the full benchmark.
\end{itemize}

\begin{figure*}[tbp]
    \centering
    \begin{subfigure}[b]{.32\linewidth} 
        \centering
        \includegraphics[width=\linewidth]{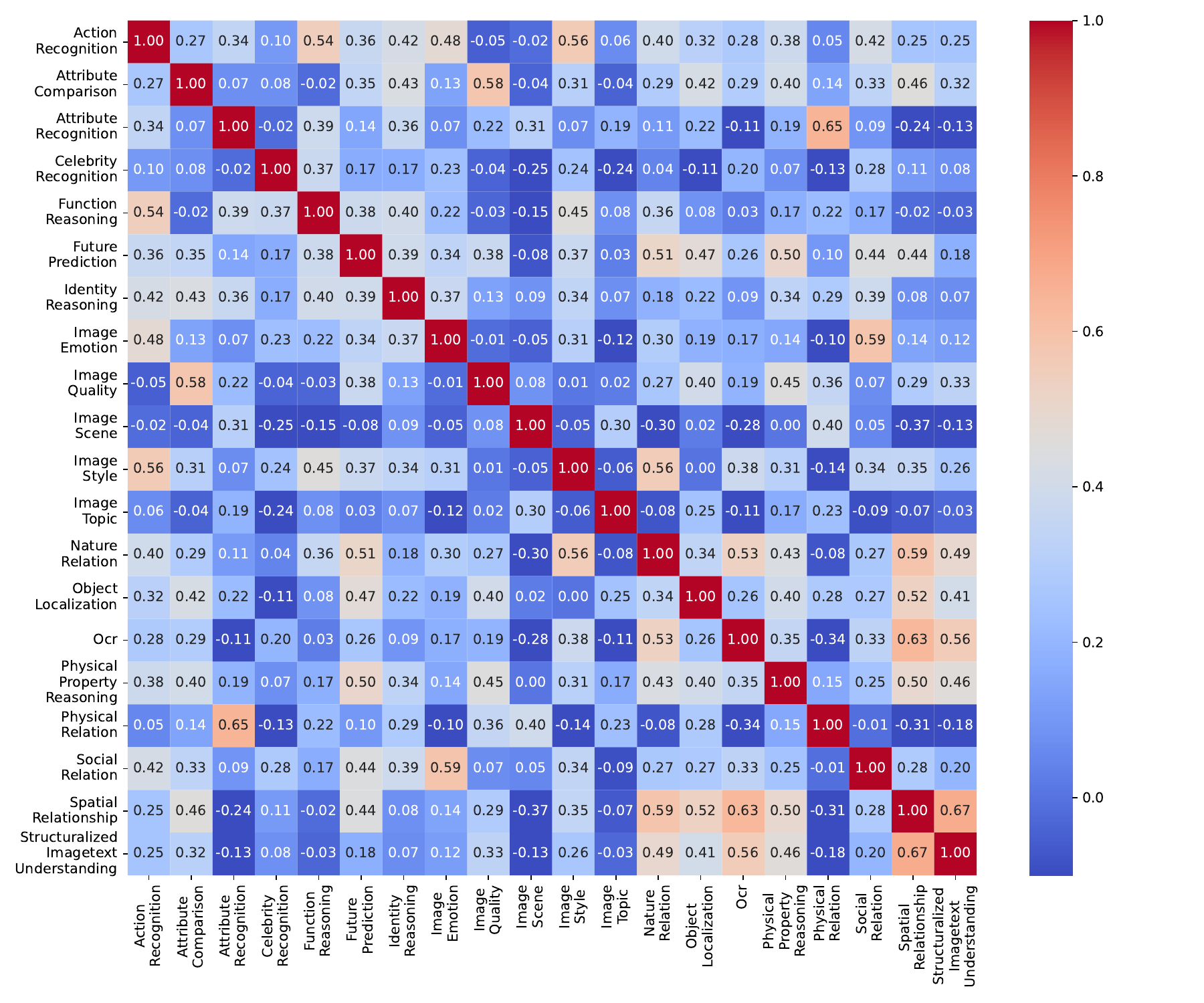}
        \caption{50$^+$ SRCC dimensions redundancy.}
        \label{fig:MMBench_SRCC_50_heat}
    \end{subfigure}
    \hfill
    \begin{subfigure}[b]{.32\linewidth} 
        \centering
        \includegraphics[width=\linewidth]{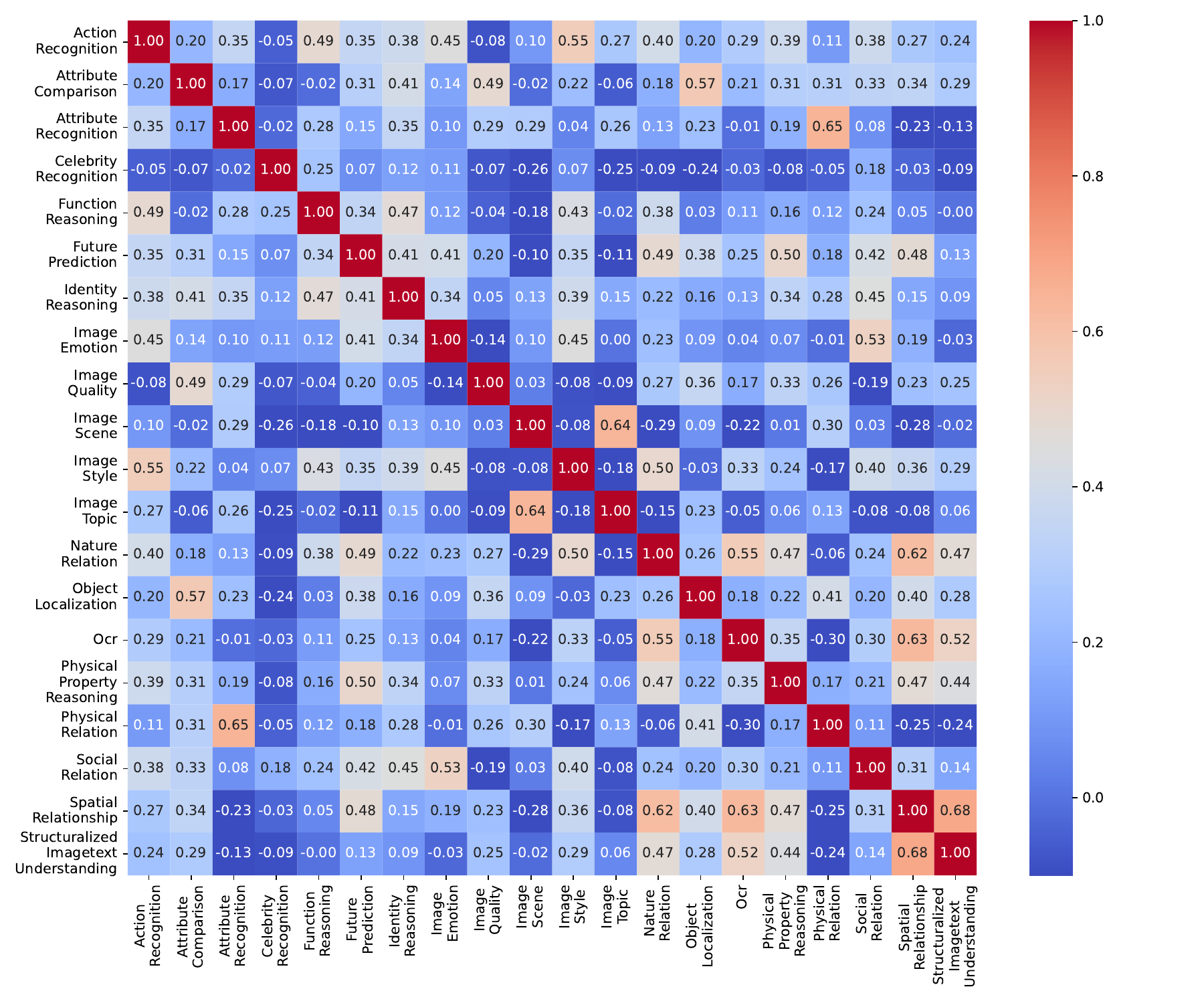}
        \caption{50$^+$ PLCC dimensions redundancy.}
        \label{fig:MMBench_PLCC_50_heat}
    \end{subfigure}
    \hfill
    \begin{subfigure}[b]{.32\linewidth} 
        \centering
        \includegraphics[width=\linewidth]{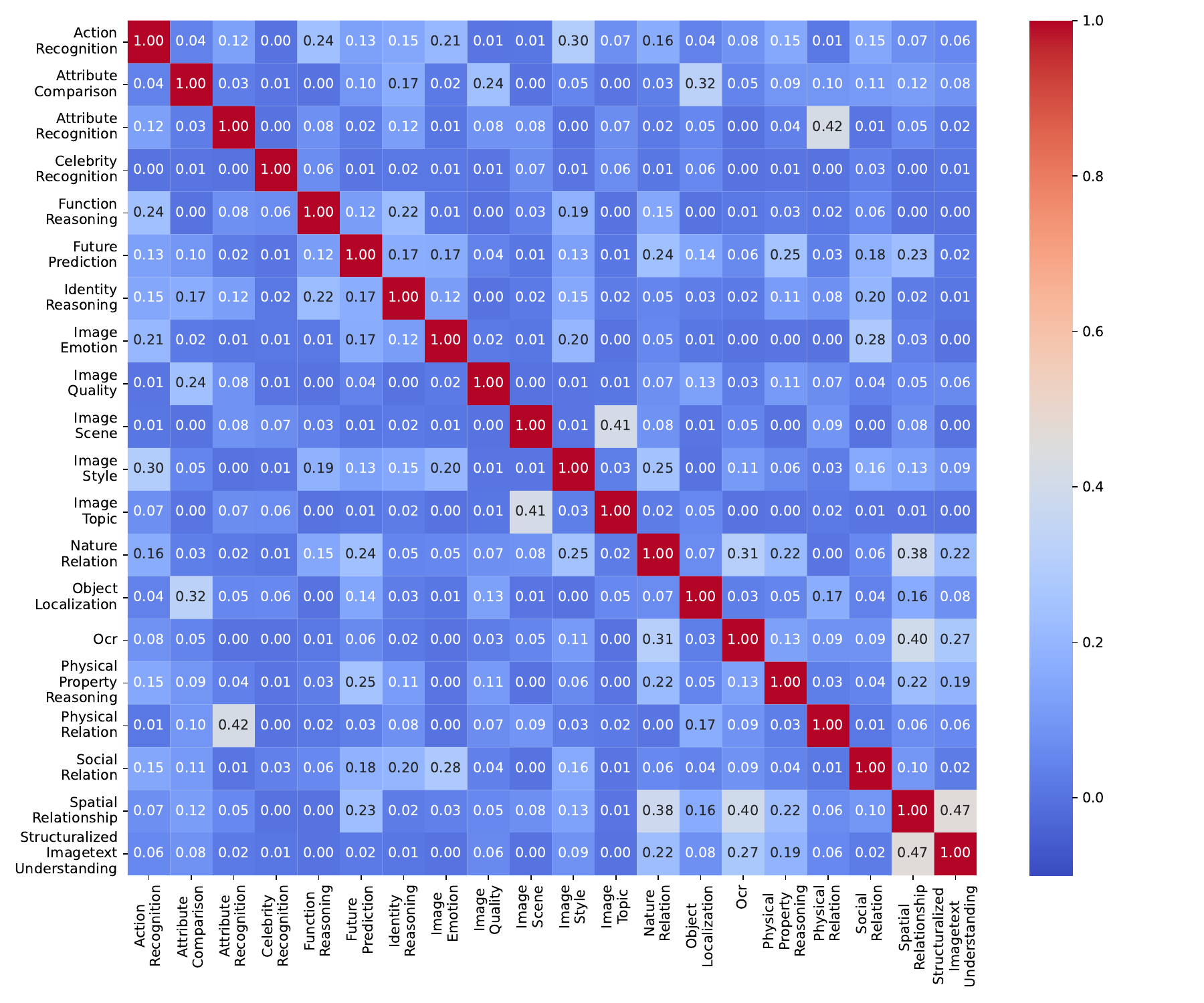}
        \caption{50$^+$ R2 dimensions redundancy.}
        \label{fig:MMBench_R2_50_heat}
    \end{subfigure}
    \begin{subfigure}[b]{.32\linewidth}  
        \centering
        \includegraphics[width=\linewidth]{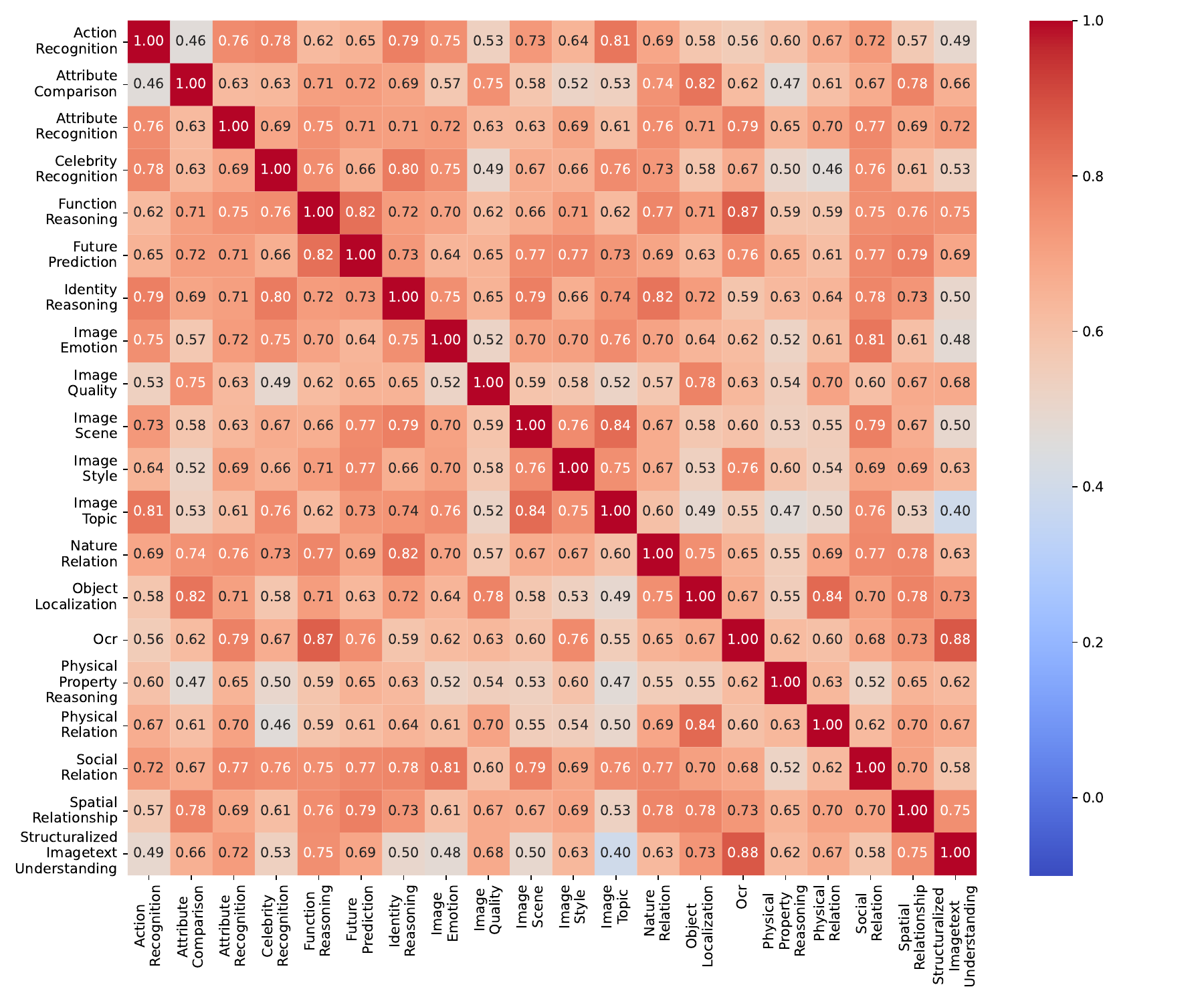}
        \caption{50$^-$ SRCC dimensions redundancy.}
        \label{fig:MMBench_SRCC_-50_heat}
    \end{subfigure}
    \hfill
    \begin{subfigure}[b]{.32\linewidth}  
        \centering
        \includegraphics[width=\linewidth]{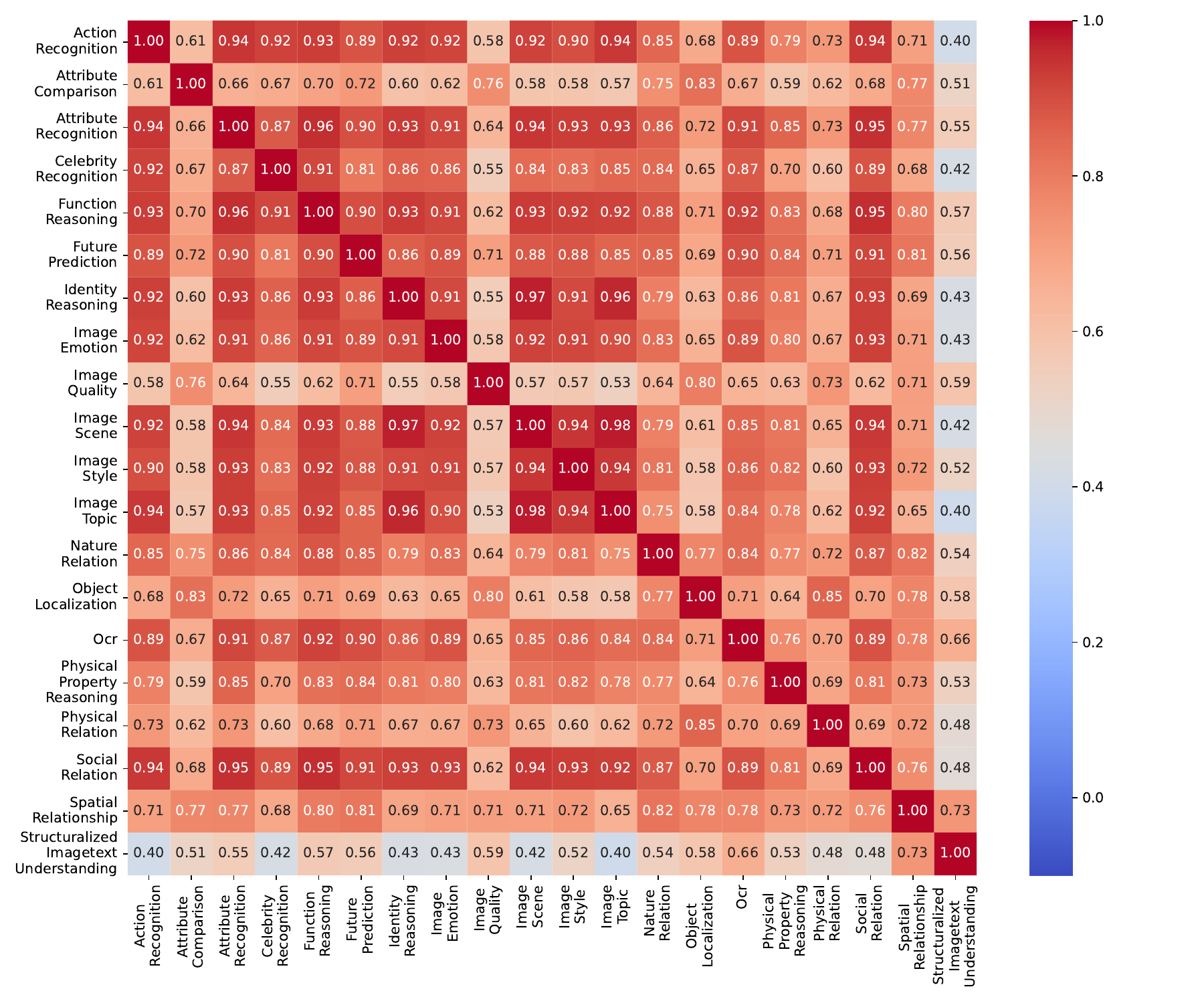}
        \caption{50$^-$ PLCC dimensions redundancy.}
        \label{fig:MMBench_PLCC_-50_heat}
    \end{subfigure}
    \hfill
    \begin{subfigure}[b]{.32\linewidth}  
        \centering
        \includegraphics[width=\linewidth]{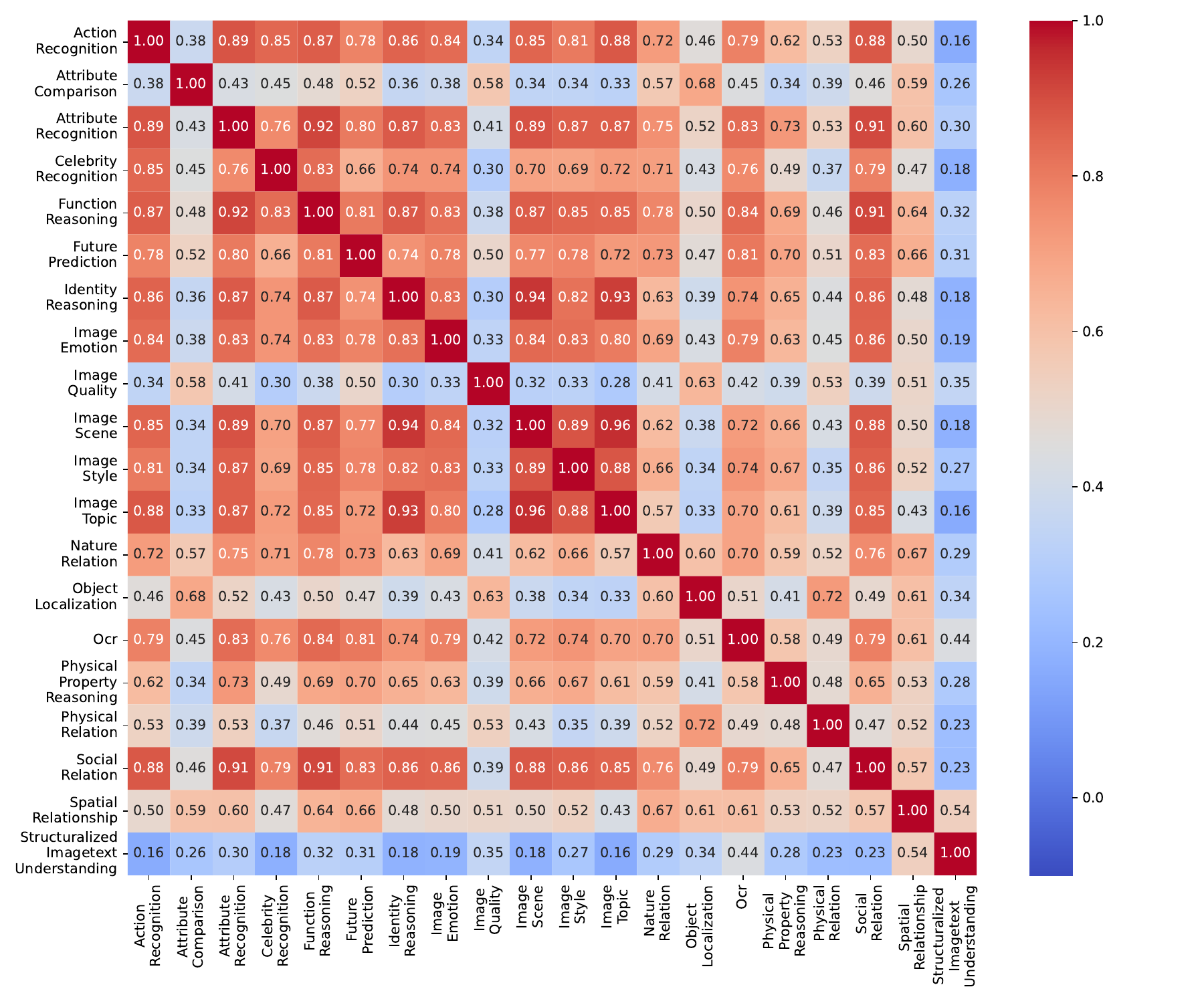}
        \caption{50$^-$ R2 dimensions redundancy.}
        \label{fig:MMBench_R2_-50_heat}
    \end{subfigure}
    \caption{Visualizations of dimensions redundancy for MMBench~\cite{liu2025mmbench} on Top-50 and Bottom-50 (\textit{marked as 50$^+$ and 50$^-$}) MLLMs respectively.  More benchmark results can be found in Appendix.~\ref{app:extra}.}
    \label{fig:MMbench_heat}
    \vspace{-10pt}
\end{figure*}

\subsection{Cross-Benchmark Redundancy}
\label{sec2.3}
Consider \( Y = \{Y_1, Y_2, \ldots, Y_l\} \), 
a collection of \( l \) benchmarks within a specific domain
(\textit{e.g.}, object hallucination, visual reasoning, visual perception). 
Let \( N \) represent the number of MLLMs evaluated across these benchmarks. 
For a given benchmark \( Y_i \), 
let \( K_i \) denote the ranking of the \( N \) MLLMs based on their performance on \( Y_i \). 
To identify key anchor benchmarks within this domain 
(an anchor benchmark can serve as a representative over multiple other benchmarks),  
we focus on benchmarks that demonstrate high redundancy with others in the domain~\cite{zohar2024apollo}.
We define the redundancy of a benchmark \( \rho(Y_i) \) as the average rank correlation coefficient between \( K_i \) and the rankings \( K_j \) of all other benchmarks \( Y_j \) (\( j \neq i \)) in the domain. 
Formally, \( \rho(Y_i) \) is expressed as:
\begin{equation}
   \rho(Y_i) = \frac{1}{l-1} \sum_{\substack{j=1 \\ j \neq i}}^l \text{CORR}(K_i, K_j), 
\end{equation}
where \( \text{CORR}(K_i, K_j) \) is the correlation coefficient between the rankings \( K_i \) and \( K_j \). 
The interpretation of \( \rho(Y_i) \) is as follows:

\begin{itemize}[left=0pt, itemsep=0pt, topsep=0pt, parsep=0pt]
    \item A higher \( \rho(Y_i) \) indicates that benchmark \( Y_i \) exhibits strong similarity with others in the domain, 
    suggesting that it is highly representative of the domain’s capabilities or evaluation focus.
    \item Conversely, a lower \( \rho(Y_i) \) indicates that benchmark \( Y_i \) shares less overlap with others, implying that it is less redundant and may capture unique/distinct aspects of the domain, or incorporate noises that are not related to the domain.
\end{itemize}
% This metric helps in identifying benchmarks that best capture the core characteristics of the domain, facilitating the selection of a reduced yet representative subset of benchmarks.

\subsection{Correlation Metrics}
In this work, we adopt multiple metrics to describe the correlation between two set of performance numbers, 
including the Spearman Rank Correlation Coefficient (SRCC), 
the Pearson Linear Correlation Coefficient (PLCC), 
and the R² Score (R-squared Coefficient of Determination).
\begin{itemize}[left=0pt, itemsep=0pt, topsep=0pt, parsep=0pt]
\item \textbf{SRCC} is an evaluation metric that measures rank similarity, capturing how well the relative order between two rankings aligns.
\item \textbf{PLCC} quantifies linear similarity, assessing how closely the rankings follow a linear relationship.
\item \textbf{R² Score}, on the other hand, evaluates the proportion of variance explained by the ranking relationship, serving as a measure of goodness-of-fit.
\end{itemize}

\subsection{Top-K Analysis}

Considering that the performance of top-tier MLLMs often garners greater attention on benchmarks, 
we can streamline the redundancy analysis by focusing only on the top-K MLLMs with the highest overall performance on a given benchmark, 
rather than incorporating all MLLMs in the calculation. 
By selecting the top-K models, we can better target the analysis of benchmark redundancy across different performance tiers. 
This approach also simplifies the process of maintaining and updating our framework as new MLLMs are introduced.

\begin{figure*}[tbp]
    \centering
    \begin{subfigure}[b]{.32\linewidth} 
        \centering
        \includegraphics[width=\linewidth]{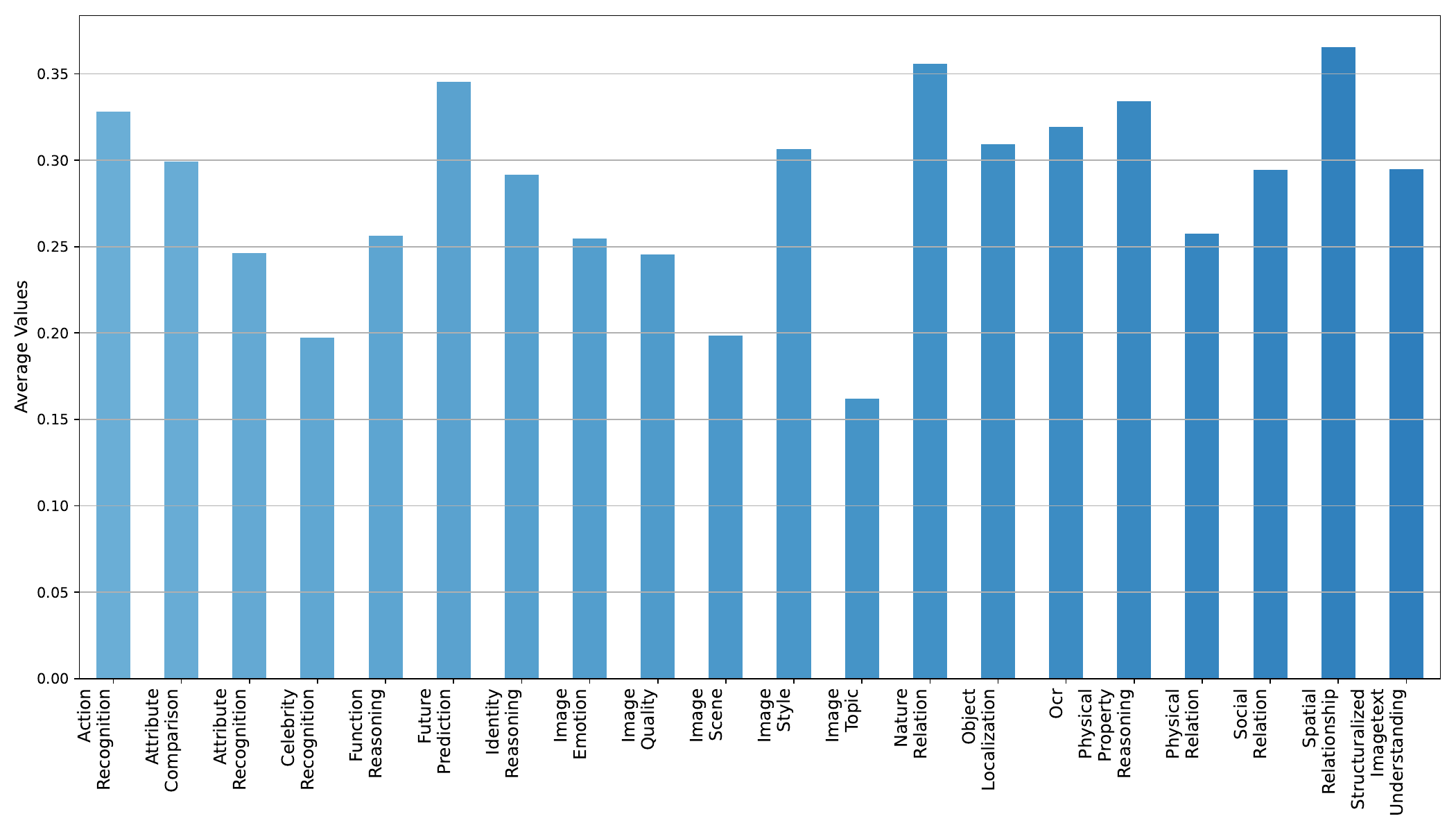}
        \caption{Top-50 SRCC redundancy.}
        \label{fig:MMBench_SRCC_50_bar}
    \end{subfigure}
    \hfill
    \begin{subfigure}[b]{.32\linewidth}  
        \centering
        \includegraphics[width=\linewidth]{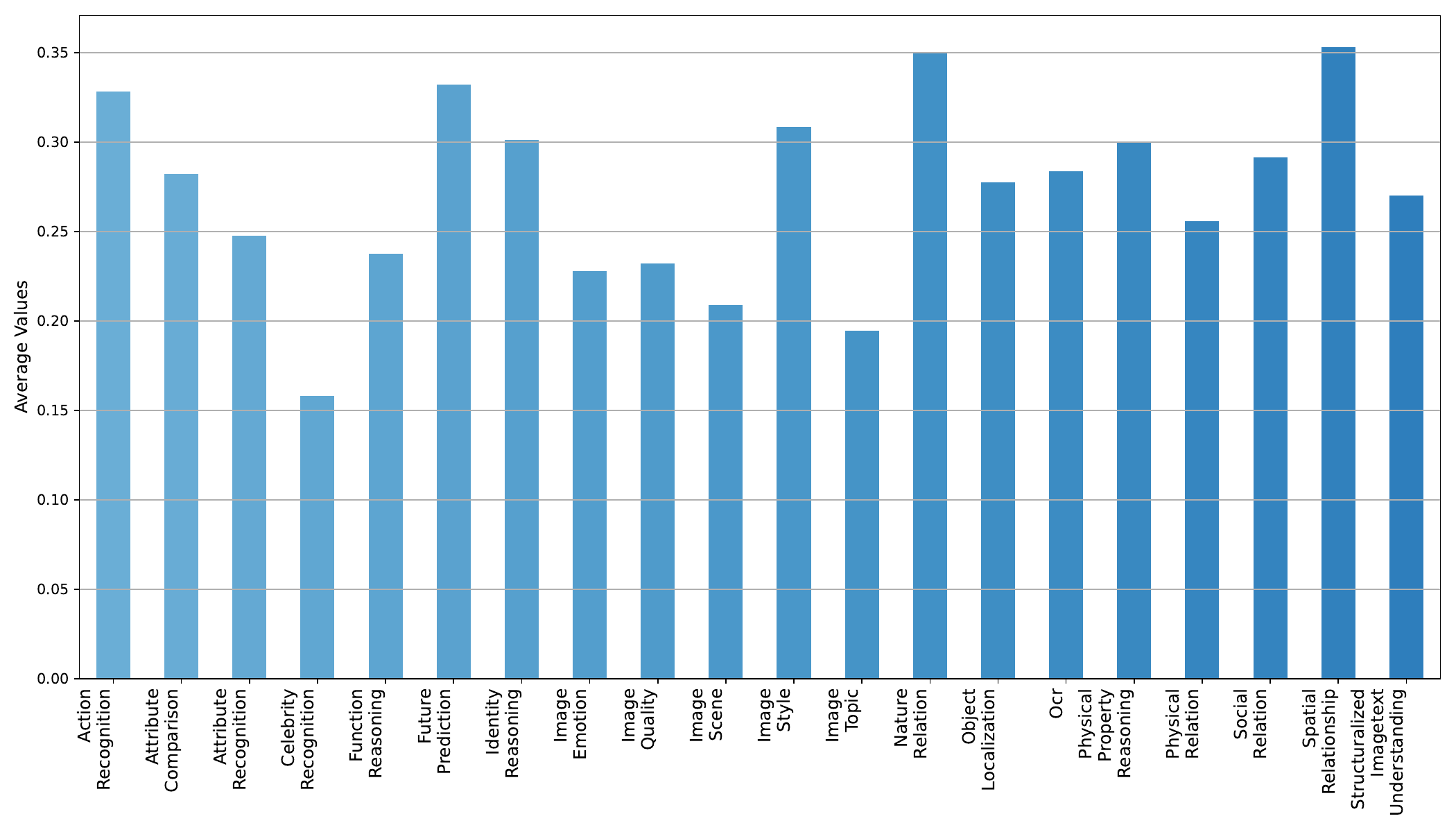}
        \caption{Top-50 PLCC redundancy.}
        \label{fig:MMBench_PLCC_50_bar}
    \end{subfigure}
    \hfill
    \begin{subfigure}[b]{.32\linewidth}  
        \centering
        \includegraphics[width=\linewidth]{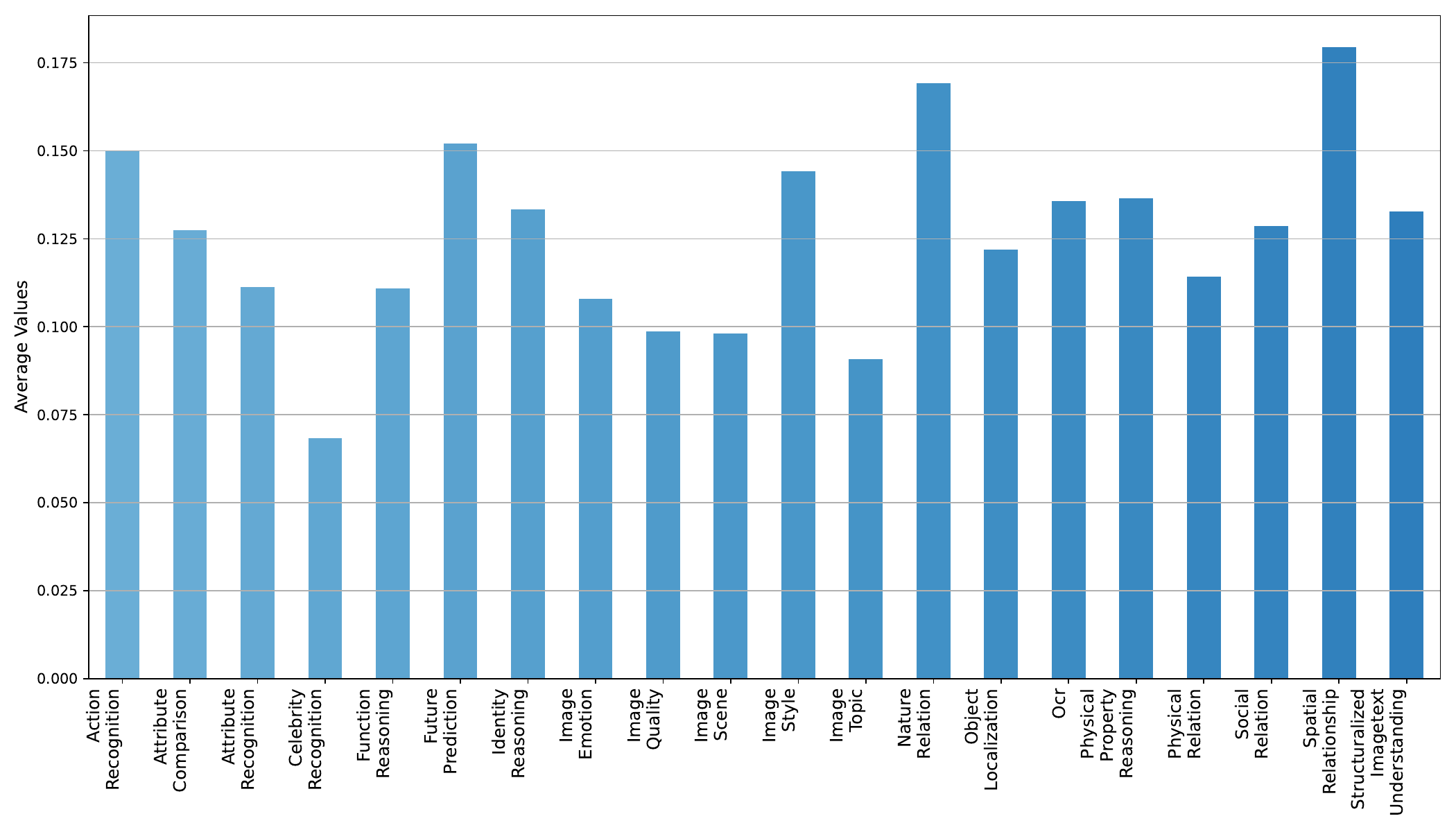}
        \caption{Top-50 R2 redundancy.}
        \label{fig:MMBench_R2_50_bar}
    \end{subfigure}
    \begin{subfigure}[b]{.32\linewidth} 
        \centering
        \includegraphics[width=\linewidth]{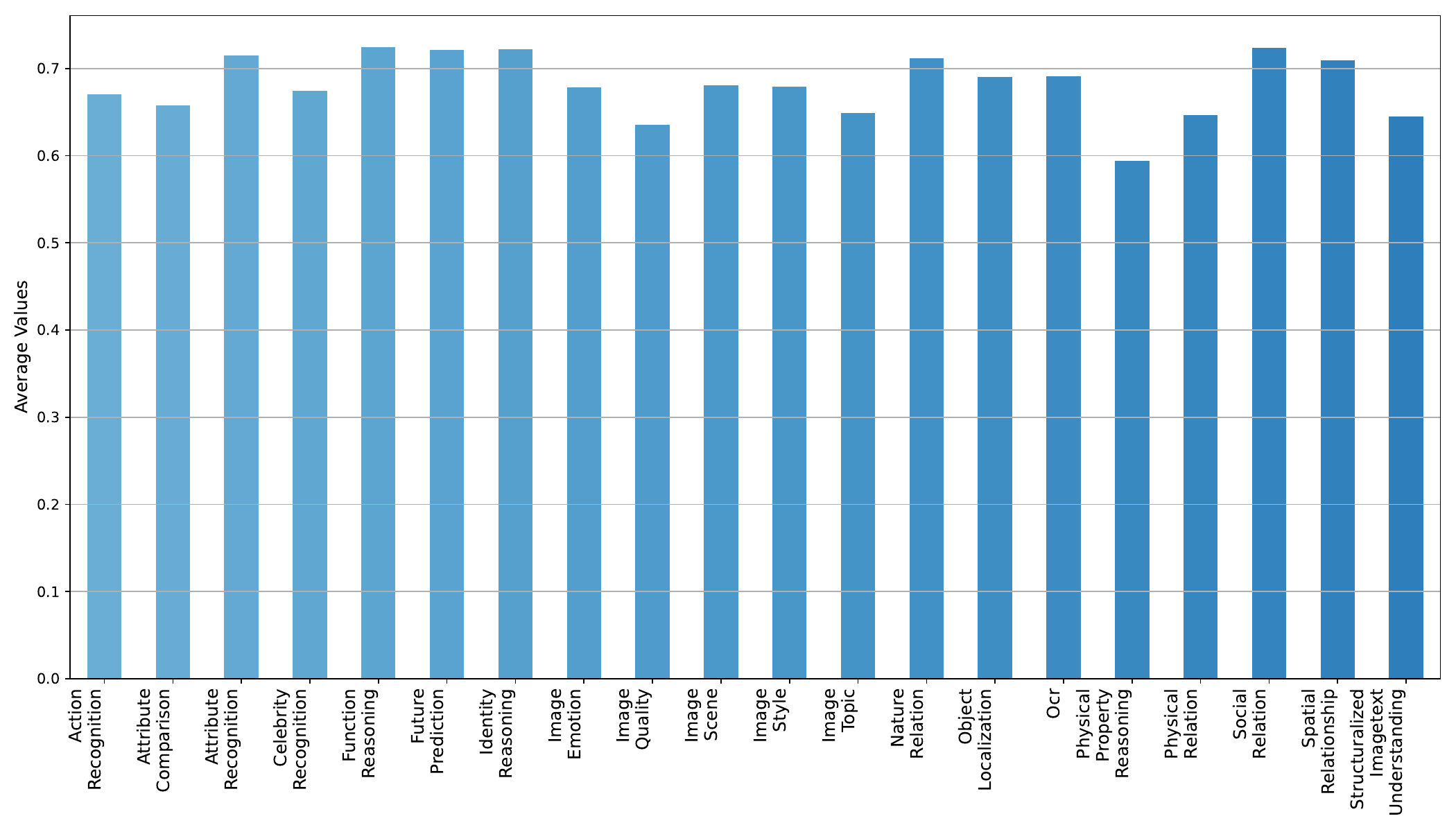}
        \caption{Bottom-50 SRCC redundancy.}
        \label{fig:MMBench_SRCC_-50_bar}
    \end{subfigure}
    \hfill
    \begin{subfigure}[b]{.32\linewidth}  
        \centering
        \includegraphics[width=\linewidth]{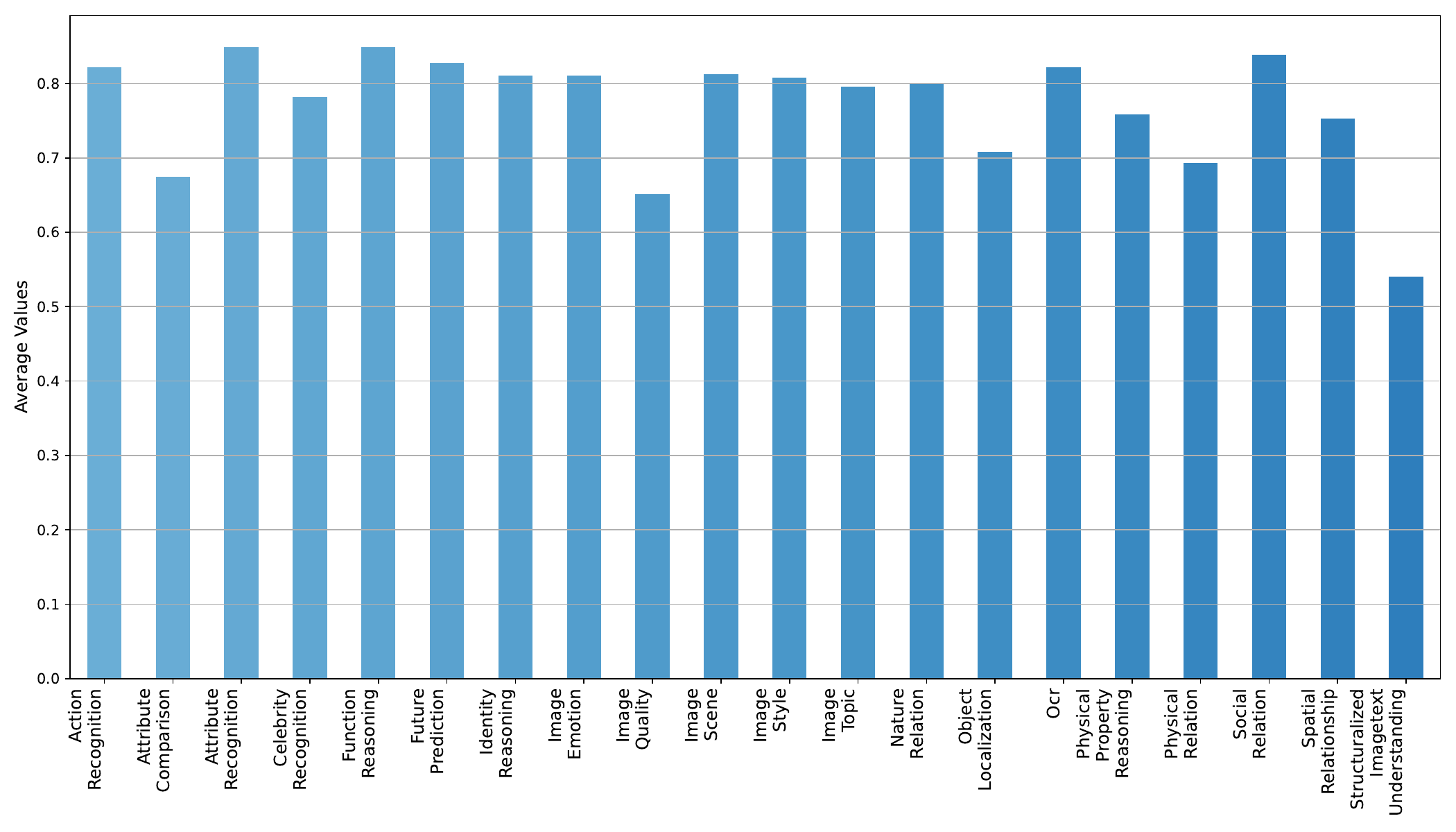}
        \caption{Bottom-50 PLCC redundancy.}
        \label{fig:MMBench_PLCC_-50_bar}
    \end{subfigure}
    \hfill
    \begin{subfigure}[b]{.32\linewidth}  
        \centering
        \includegraphics[width=\linewidth]{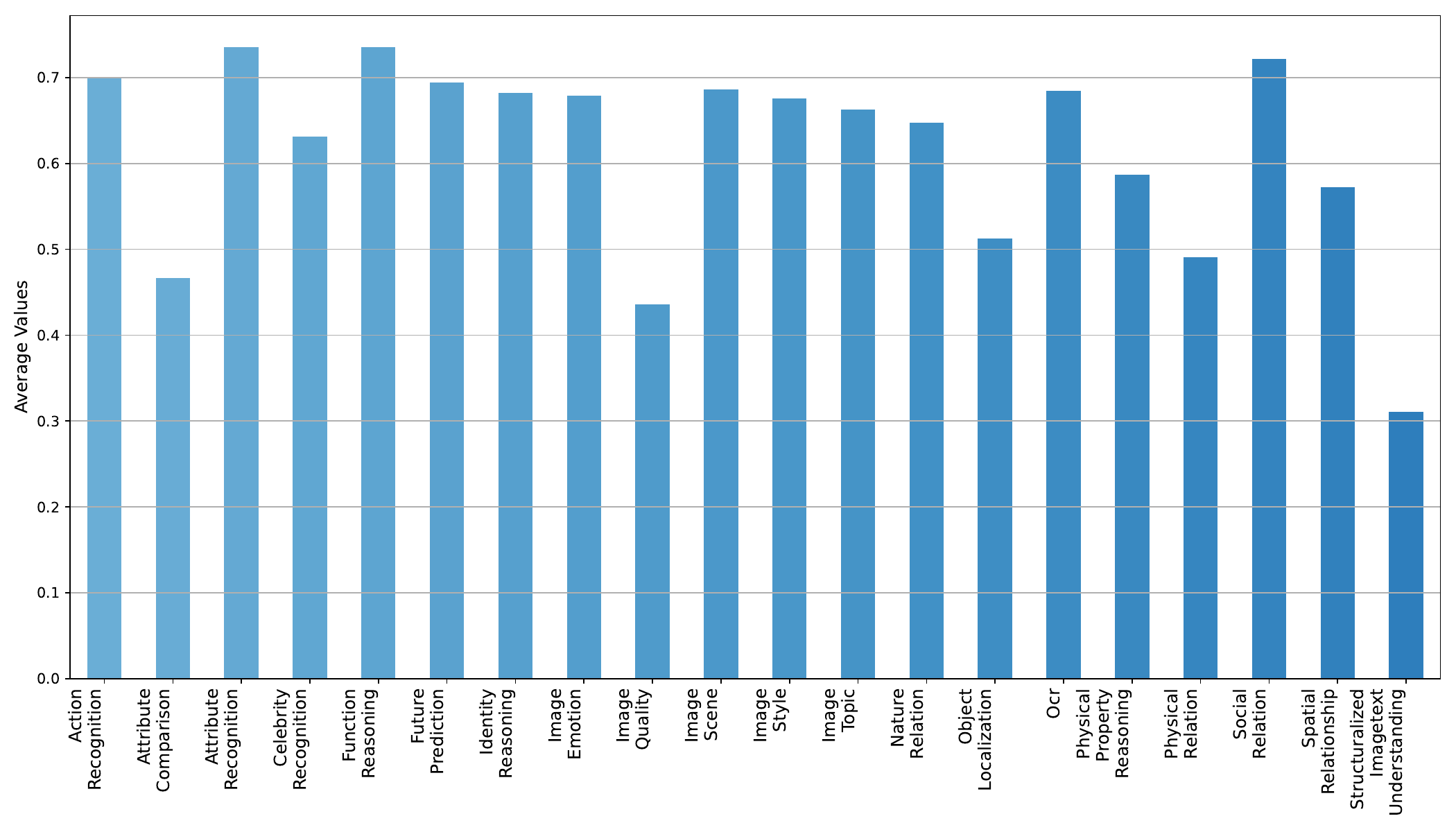}
        \caption{Bottom-50 R2 redundancy.}
        \label{fig:MMBench_R2_-50_bar}
    \end{subfigure}
    \caption{Bar plots of dimensions redundancy for MMBench~\cite{liu2025mmbench} on Top-50 and Bottom-50 MLLMs. The redundancy values are computed by averaging the redundancy of each dimension with the redundancy of all other dimensions.}
    \label{fig:MMbench_bar}
    \vspace{-10pt}
\end{figure*}

\section{Experiment \& Discussion}

We use the evaluation results of hundreds of MLLMs obtained through the VLMEvalKit~\cite{duan2024vlmevalkit} as our data source 
for conducting experiments and analysis. 
All the data sources are open-sourced and available on HuggingFace \footnote{\url{https://huggingface.co/datasets/VLMEval/OpenVLMRecords}}.

% \subsection{Data Source}

% All the benchmark performance results come from the OpenCompass platform, which includes over xxx popular benchmarks and xxx MLLMs. 

\subsection{Exploring Dimension Redundancy}

To comprehensively demonstrate the application of our redundancy framework in MLLM benchmarks, 
we conduct a detailed case study using the widely adopted and dimensionally diverse MMBench benchmark (v1.1)~\cite{liu2025mmbench}. 
We categorize the MLLMs into two groups, Top-50 and Bottom-50, based on their overall performance in MMBench. 
This categorization enables us to highlight the differences in redundancy exhibited by MMBench when evaluating MLLMs with varying levels of capability. 
The dimension redundancy results are illustrated in \cref{fig:MMbench_heat} and \cref{fig:MMbench_bar}, 
from which we derived several interesting insights.

\noindent
\textbf{Top-50 Redundancy. }
\cref{fig:MMBench_SRCC_50_heat,fig:MMBench_PLCC_50_heat} visually illustrate the redundancy of SRCC and PLCC across various sub-dimensions,
allowing for a quick analysis of which dimensions exhibit high correlations. 
For example, the tasks \textbf{Image Emotion} and \textbf{Social Relation} display strong redundancy, suggesting a significant overlap in the skills they assess.
Similarly, \textbf{Structuralized Image-Text Understanding} demonstrates notable redundancy with several other dimensions, such as \textbf{Spatial Relationship}, \textbf{Physical Property Reasoning}, \textbf{OCR}, and \textbf{Nature Relation}, 
indicating that these tasks collectively represent the diverse abilities required to perform \textbf{Structuralized Image-Text Understanding}. 
In contrast, \textbf{Image Topic} and \textbf{Image Scene} exhibit relatively low redundancy with other dimensions, as shown in \cref{fig:MMBench_SRCC_50_bar,fig:MMBench_PLCC_50_bar,fig:MMBench_R2_50_bar}. 
This could arise from the inherent complexity of assessing the overall topic and scene of an image, 
which is often less correlated with evaluating specific attributes or relationships.
For instance, strong performance in recognizing individual attributes does not necessarily imply a comprehensive understanding of the overall topic or scene. 
However, \cref{fig:MMBench_PLCC_50_heat} reveals that these two dimensions exhibit redundancy in terms of PLCC, suggesting potential overlaps within certain contexts.
Another interesting insight arises from \textbf{Celebrity Recognition}, 
a knowledge-based task that remains relatively independent of other dimensions, 
which primarily measure perceptual abilities. 
As a result, it consistently exhibits significantly lower redundancy across SRCC, PLCC, and R². 
Conversely, high levels of redundancy are observed for Nature Relation and Spatial Relationship, as shown in \cref{fig:MMBench_SRCC_50_bar,fig:MMBench_PLCC_50_bar,fig:MMBench_R2_50_bar}. 
This is attributed to the fact that these two dimensions serve as fundamental skills required by numerous other tasks, making their overlap a cornerstone of the broader evaluation framework.

\begin{figure*}[htbp]
    \centering
    \begin{subfigure}[b]{.48\linewidth}  % Adjust width to fit side by side
        \centering
        \includegraphics[width=\linewidth]{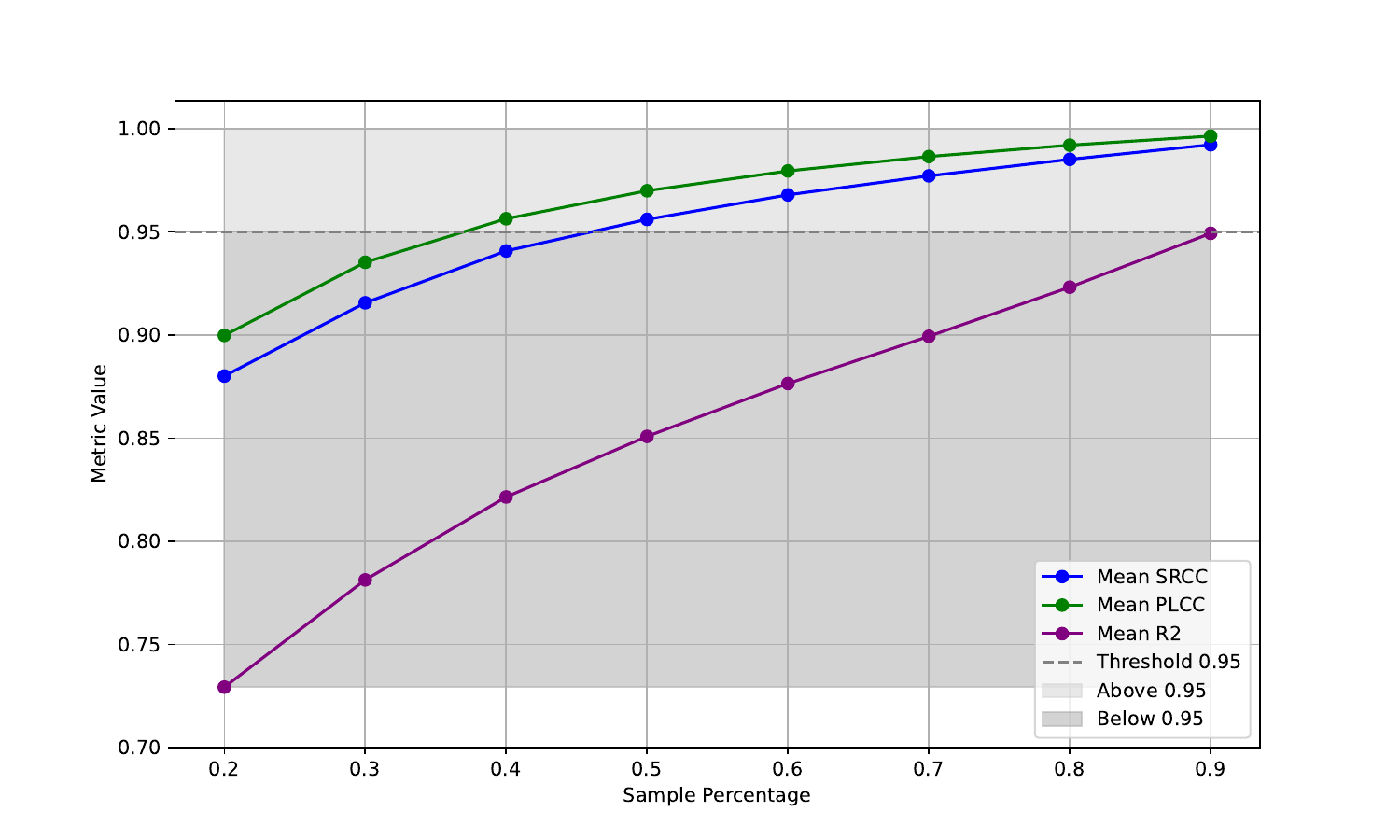}
        \caption{Instances redundancy with Top-50 MLLMs.}
        \label{fig:instances_top50}
    \end{subfigure}
    \hfill
    \begin{subfigure}[b]{.48\linewidth}  % Ensure both subfigures are the same width
        \centering
        \includegraphics[width=\linewidth]{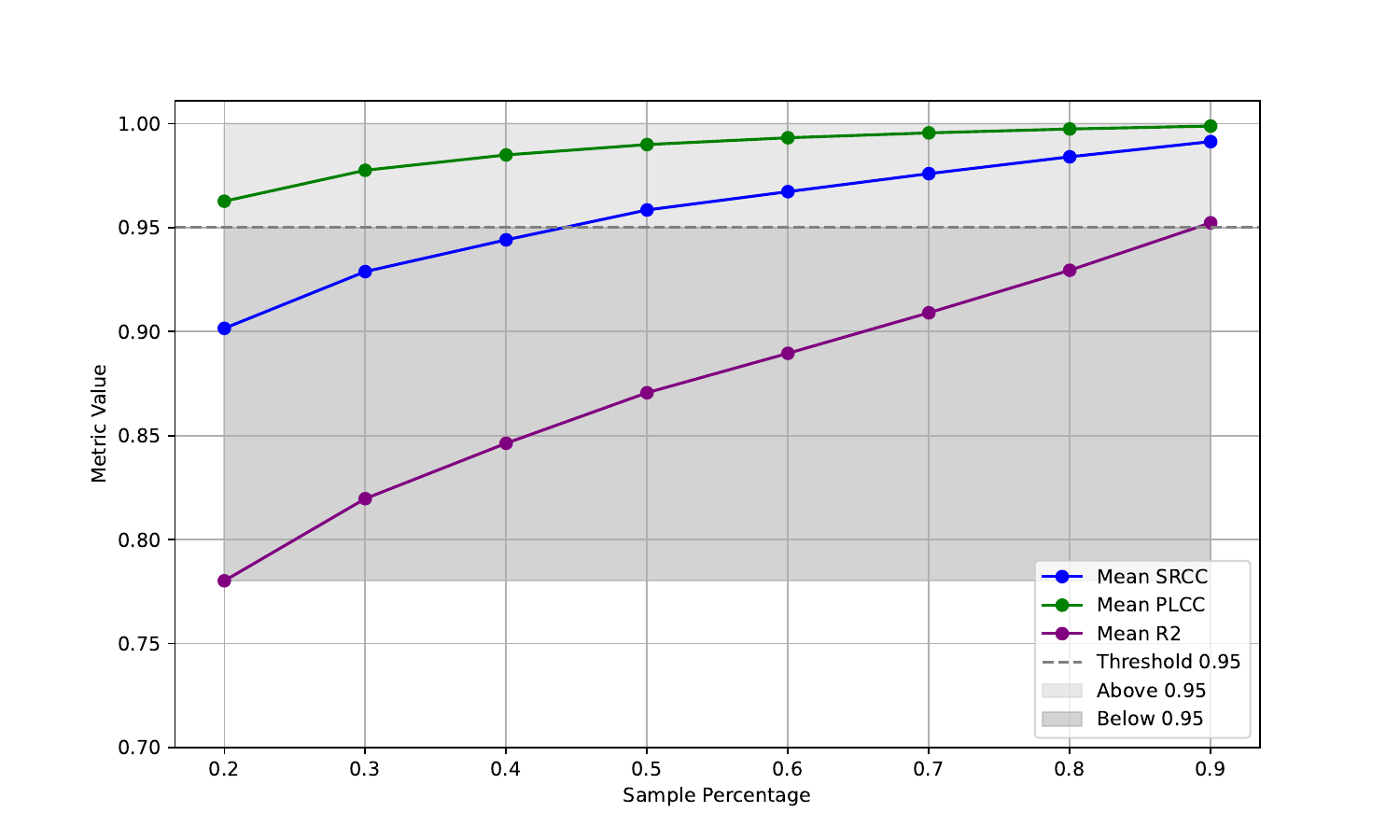}
        \caption{Instances redundancy with Bottom-50 MLLMs.}
        \label{fig:instances_back50}
    \end{subfigure}
    \caption{Visualizations of average instance redundancy for (a) Top-50 MLLMs and (b) Bottom-50 MLLMs across 18 LMM benchmarks (A-Bench~\cite{zhang2024bench}, AI2D~\cite{kembhavi2016diagram}, BLINK~\cite{fu2025blink},  HallusionBench~\cite{guan2023hallusionbench}, MMBench~\cite{liu2025mmbench}, MMMU~\cite{yue2024mmmu}, MME~\cite{fu2024mmecomprehensiveevaluationbenchmark}, MMStar~\cite{chen2024mmstar}, MMT~\cite{ying2024mmt}, MMVet~\cite{yu2023mmvet}, OCRBench~\cite{liu2023ocrbench}, Q-Bench~\cite{wu2023qbench,zhang2024q+}, R-Bench-Dis~\cite{li2024rbench}, RealWorldQA~\cite{RealWorldQA}, ScienceQA~\cite{lu2022learn}, SeedBench\_IMG~\cite{li2023seedimg}, SeedBench2\_Plus~\cite{li2024seed2plus}). Notably, each data point represents the average of 100 sampling iterations to mitigate the impact of randomness. }
    \label{fig:instances_overall}
    \vspace{-0.2cm}
\end{figure*}

% 2) In contrast, simpler foundational tasks such as \textit{Spatial Relationship}, \textit{Attribute Comparison}, and \textit{Nature Relation} cover a broader range of capabilities and are more closely related to other tasks, leading to higher redundancy. These dimensions often require general-purpose reasoning skills that overlap significantly across tasks.
% 3) Tasks like \textit{Action Recognition}, \textit{Celebrity Recognition}, and \textit{Social Relation} require more specialized knowledge or expertise that may not directly correlate with general capabilities. As a result, these tasks typically exhibit lower redundancy.
% Overall, the analysis suggests that simpler tasks with broader coverage tend to exhibit higher scores and greater redundancy. Conversely, more complex or knowledge-intensive tasks demonstrate lower redundancy, highlighting their distinct and specialized evaluation focus.

\noindent
\textbf{Bottom-50 Redundancy. }
The results for the Bottom-50 redundancy, as shown in \cref{fig:MMBench_SRCC_-50_bar,fig:MMBench_PLCC_-50_bar,fig:MMBench_R2_-50_bar}, 
reveal a striking trend where nearly all dimensions exhibit significantly higher redundancy compared to the Top-50 redundancy. 
Specifically, most dimension pairs achieve SRCC and PLCC scores exceeding 0.6
(\cref{fig:MMBench_SRCC_-50_bar,fig:MMBench_PLCC_-50_bar}), 
leading to an interesting conclusion:
\textbf{the dimensions appear to be more redundant for Bottom-50 MLLMs than for Top-50 MLLMs}.
This phenomenon can primarily be attributed to the fact that Bottom-50 MLLMs generally underperform across all capabilities. 
For these models, as their foundational abilities improve, incremental enhancements in one dimension often drive simultaneous improvements across others. 
This results in high consistency in performance rankings across dimensions, thereby causing relatively high dimensional redundancy.
In contrast, the Top-50 MLLMs have already achieved relatively strong foundational capabilities. 
Consequently, more complex tasks across different dimensions introduce greater variability, allowing for more differentiation between performance in those dimensions. 
This leads to noticeably lower levels of redundancy for the Top-50 models. 
These findings emphasize the importance of carefully selecting the MLLMs included in redundancy analysis. 
Specifically, avoiding models with universally poor performance is crucial to ensure that the evaluation yields meaningful and accurate insights.

\begin{figure*}[tbp]
    \centering
    \begin{subfigure}[b]{.32\linewidth}  % Adjust width to fit side by side
        \centering
        \includegraphics[width=\linewidth]{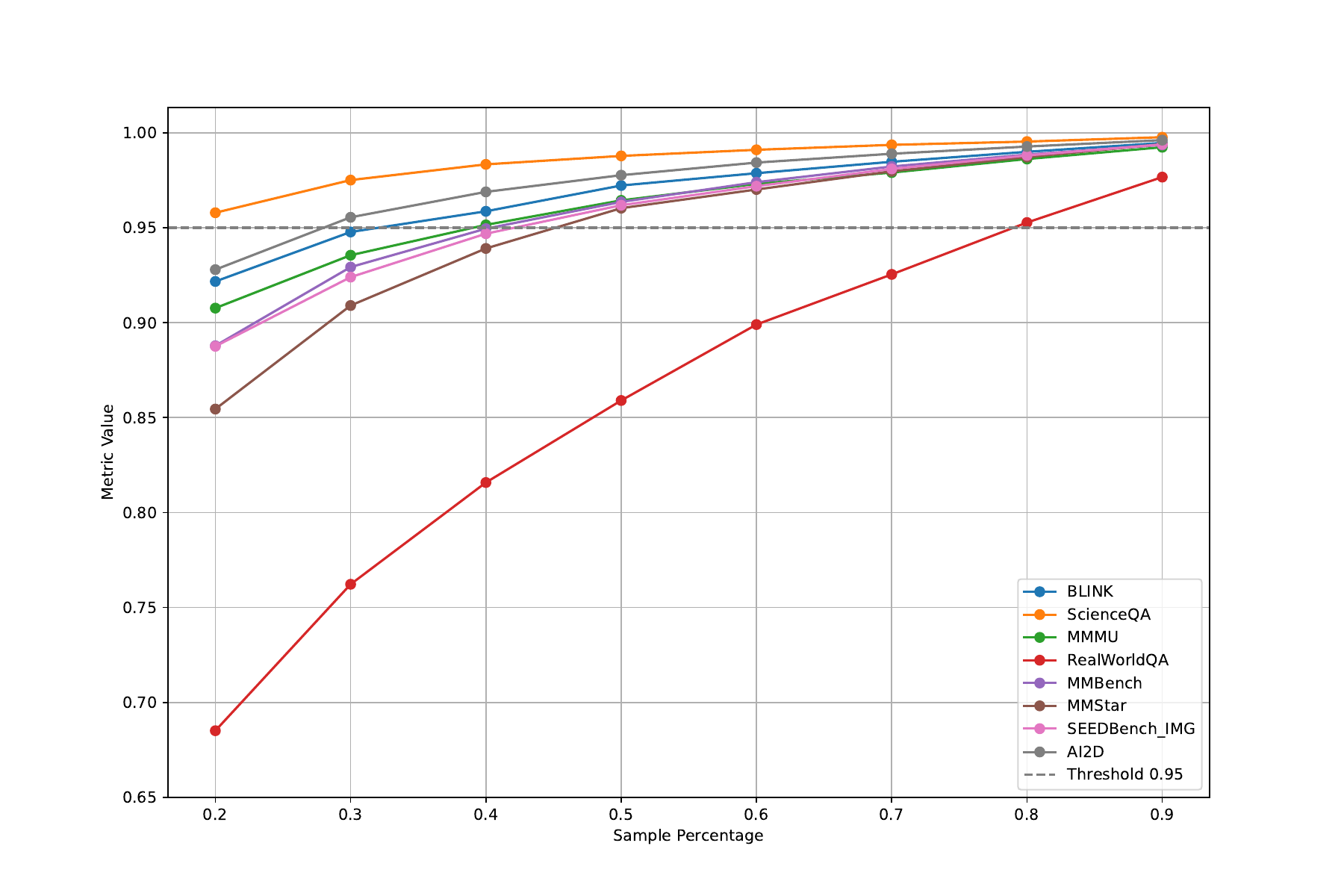}
        \caption{Top-50 SRCC redundancy.}
        \label{fig:instances_50_SRCC}
    \end{subfigure}
    \hfill
    \begin{subfigure}[b]{.32\linewidth}  % Ensure both subfigures are the same width
        \centering
        \includegraphics[width=\linewidth]{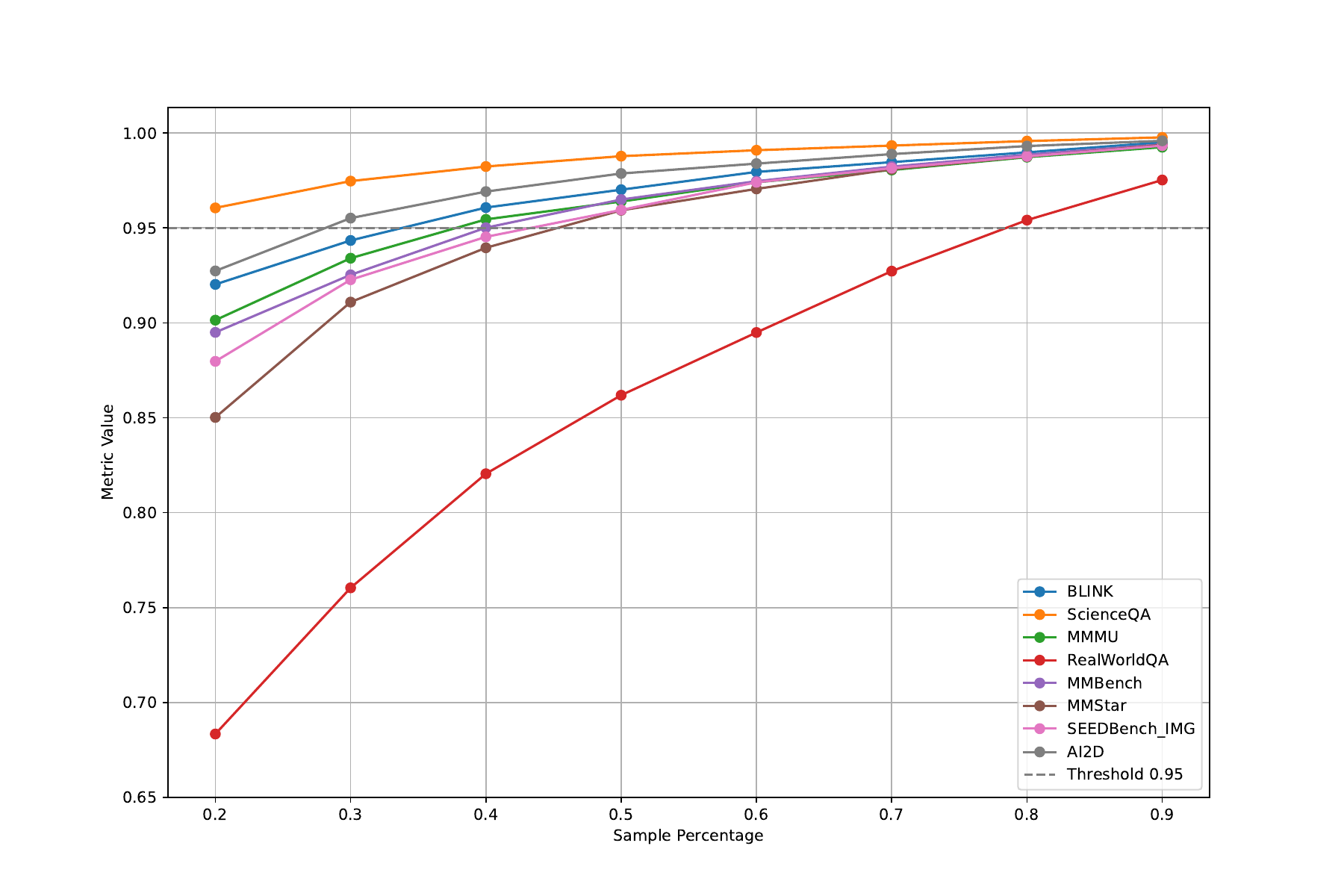}
        \caption{Top-50 PLCC redundancy.}
        \label{fig:instances_50_PLCC}
    \end{subfigure}
    \hfill
    \begin{subfigure}[b]{.32\linewidth}  % Adjust width to fit side by side
        \centering
        \includegraphics[width=\linewidth]{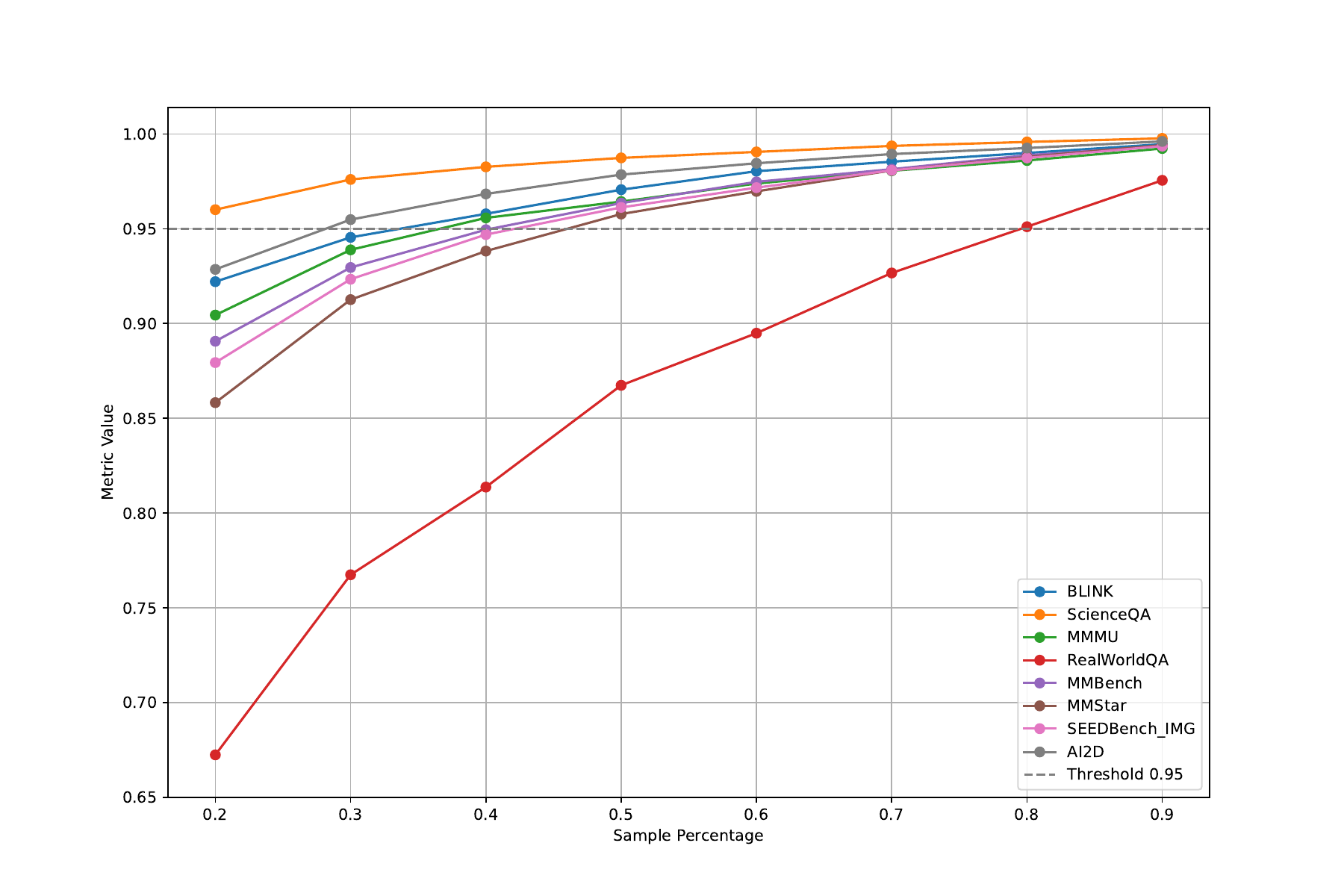}
        \caption{Top-50 R2 redundancy.}
        \label{fig:instances_50_R2}
    \end{subfigure}
    \hfill
    \begin{subfigure}[b]{.32\linewidth}  % Ensure both subfigures are the same width
        \centering
        \includegraphics[width=\linewidth]{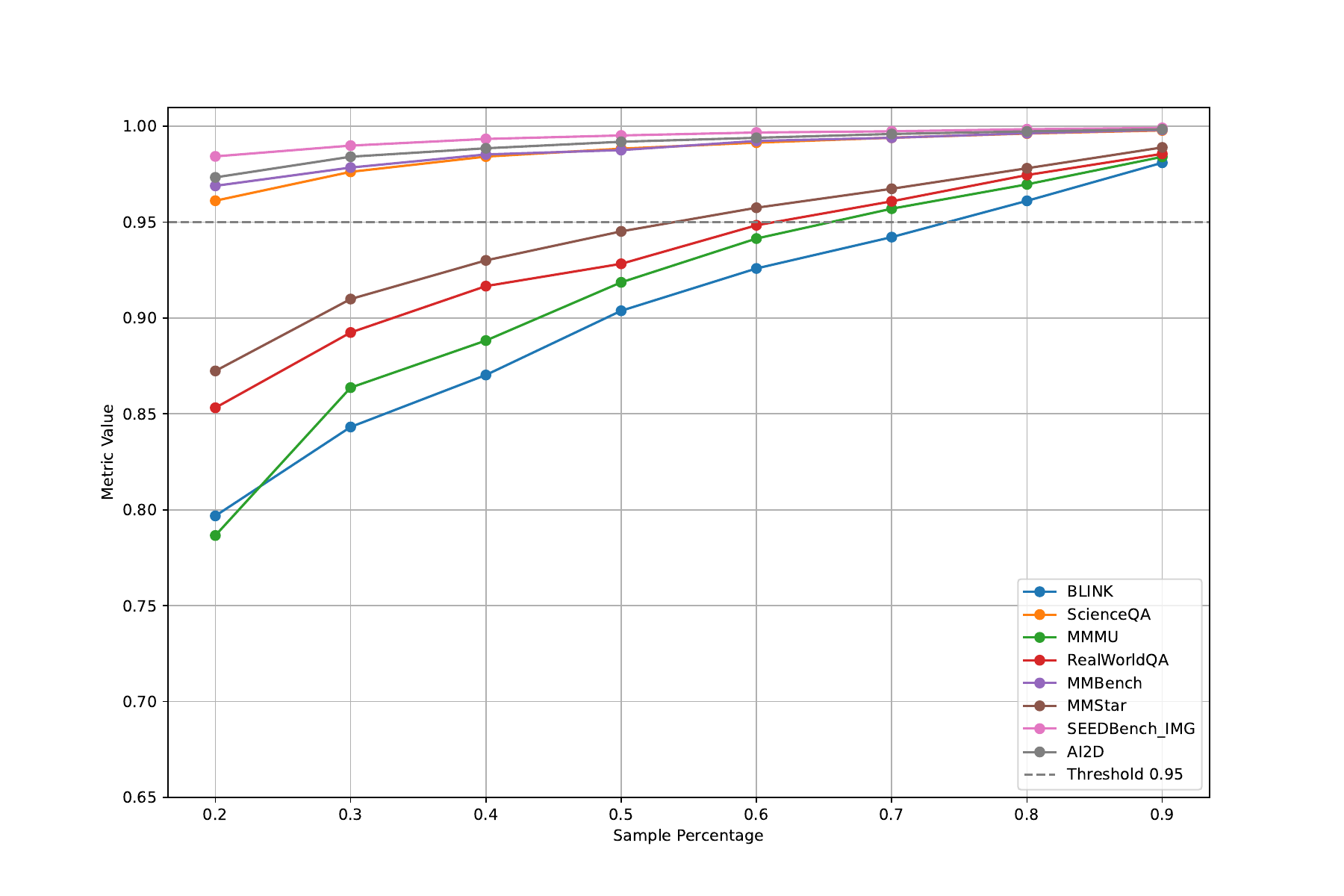}
        \caption{Bottom-50 SRCC redundancy.}
        \label{fig:instances_-50_SRCC}
    \end{subfigure}
    \hfill
    \begin{subfigure}[b]{.32\linewidth}  % Adjust width to fit side by side
        \centering
        \includegraphics[width=\linewidth]{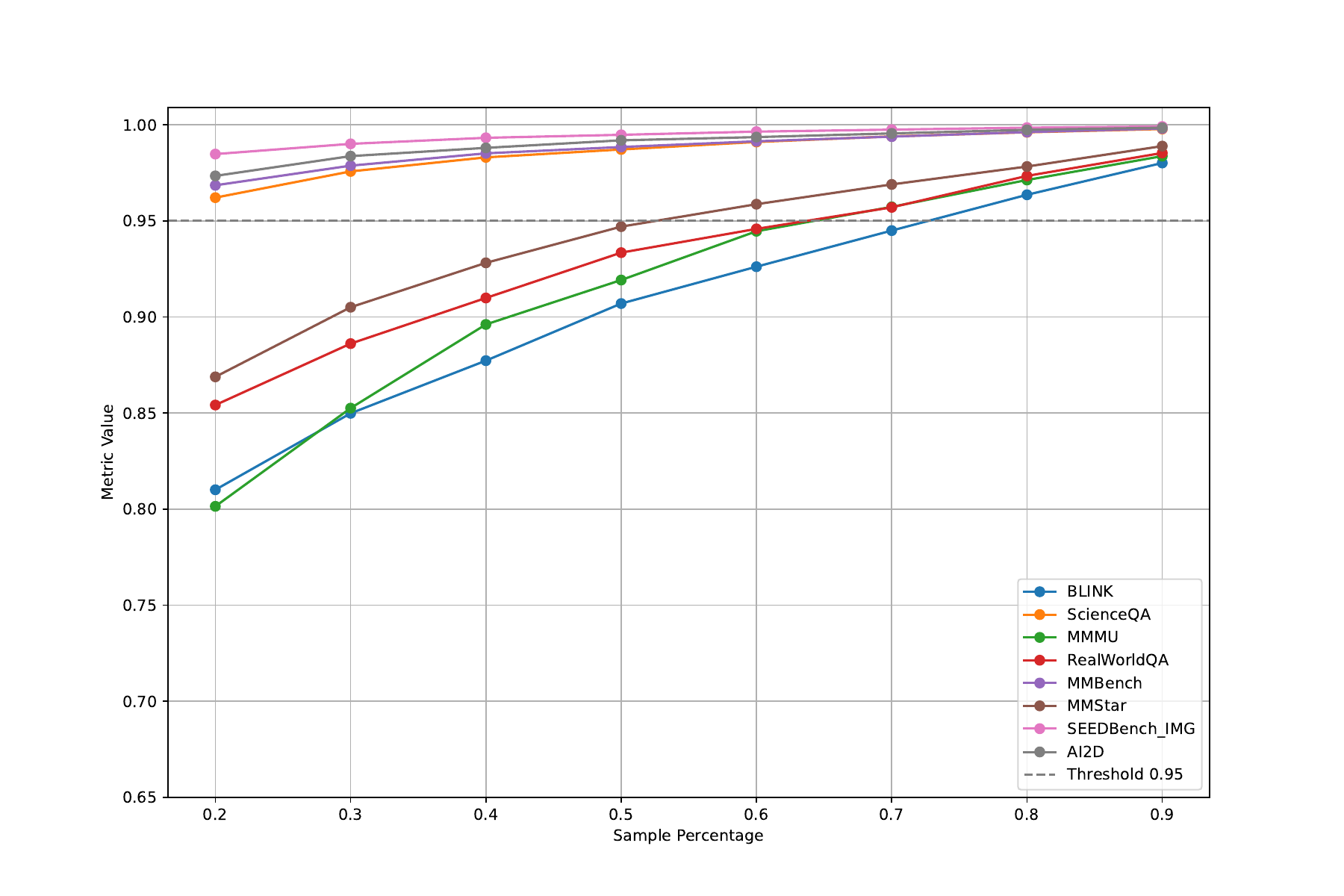}
        \caption{Bottom-50 PLCC redundancy.}
        \label{fig:instances_-50_PLCC}
    \end{subfigure}
    \hfill
    \begin{subfigure}[b]{.32\linewidth}  % Ensure both subfigures are the same width
        \centering
        \includegraphics[width=\linewidth]{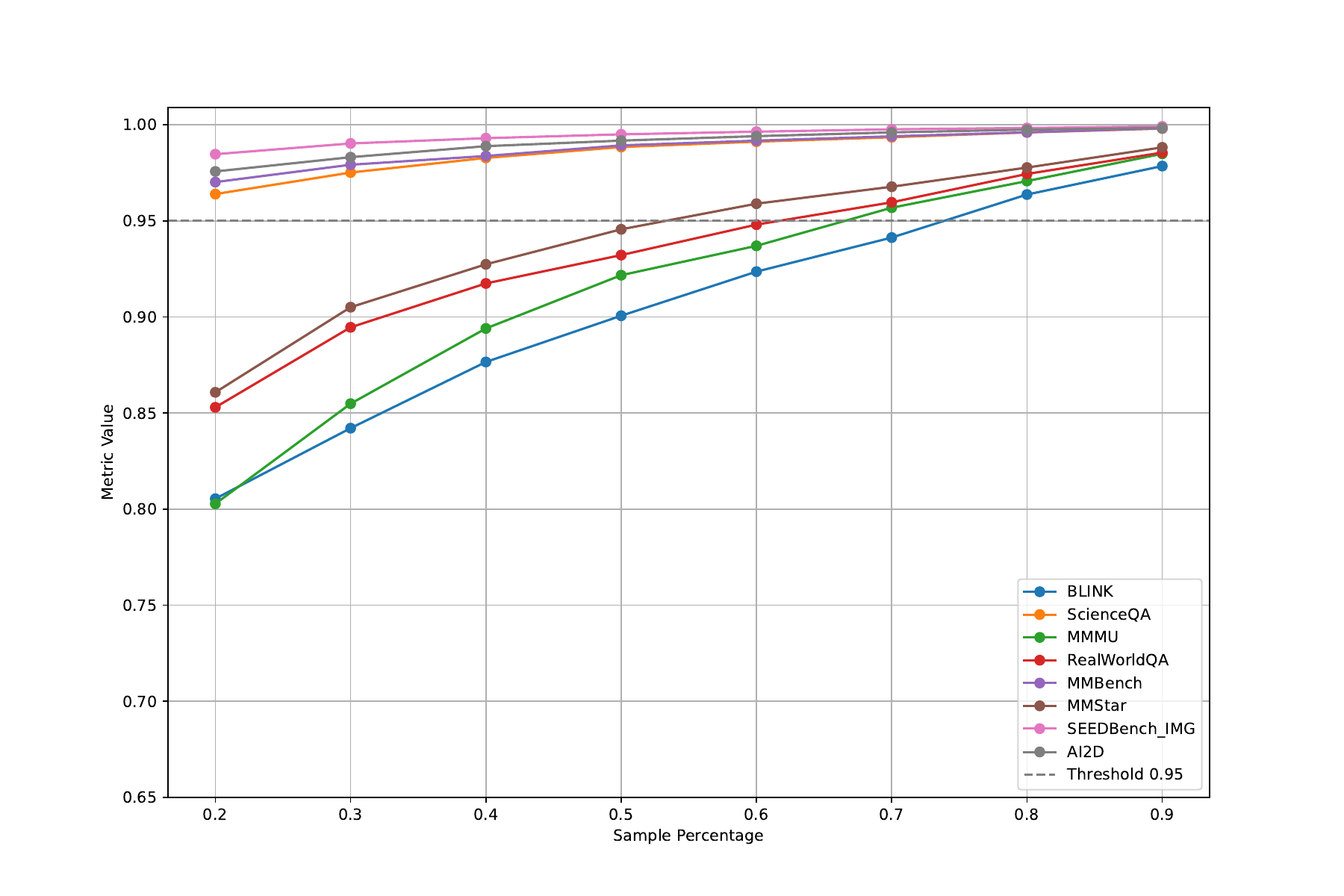}
        \caption{Bottom-50 R2 redundancy.}
        \label{fig:instances_}
    \end{subfigure}
    \caption{Benchmark-specific instance redundancy for (a) Top-50 MLLMs and (b) Bottom-50 MLLMs. The benchmarks include BLINK~\cite{fu2025blink}, ScienceQA~\cite{lu2022learn}, MMMU~\cite{yue2024mmmu}, RealWorldQA~\cite{RealWorldQA}, MMBench~\cite{liu2025mmbench}, MMStar~\cite{chen2024mmstar}, SeedBench\_IMG~\cite{li2023seedimg}, and AI2D~\cite{kembhavi2016diagram}. The selection of the Top-50 and Bottom-50 MLLMs is based on the corresponding benchmark.}
    \label{fig:instances_bench}
    \vspace{-0.2cm}
\end{figure*}

\subsection{Exploration Instance Redundancy}

We include the evaluation results from 18 publicly  available benchmarks in VLMEvalKit~\cite{duan2024vlmevalkit} in our experiments, 
with the average performance across benchmarks presented in \cref{fig:instances_overall}. 
We adopt a similarity threshold of 0.95 for partitioning\footnote{
Ranks with SRCC and PLCC coefficients exceeding 0.95 are considered nearly identical, with only marginal differences in very few cases \cite{hauke2011comparison}. }, 
This leads to an intriguing conclusion: 
\textbf{a majority of existing MLLM benchmarks exhibit significant redundancy in their instances when ranking both Top-50 and Bottom-50 MLLMs, with at least 50\% of the instances being redundant. } 
This indicates that many benchmarks could reduce their instance counts by half without significantly affecting the ranking of MLLMs being tested.
The R² score provides further insight, as it measures how effectively the final performance of MLLMs can be predicted using sampled instances. 
Compared to ensuring accurate ranking, achieving high accuracy in predicting the absolute performance of MLLMs requires a much larger number of instances. 
For example, both Top-50 and Bottom-50 MLLMs require over 90\% of the instances to achieve an R² score greater than 0.95. 
This distinction highlights that fewer instances are sufficient for reliable ranking than for precise performance prediction.

We also compare redundancy tendencies between Top-50 and Bottom-50 MLLMs, 
as shown in \cref{fig:instances_top50,fig:instances_back50}. 
Notably, at the same 0.95 threshold for SRCC and PLCC, 
Bottom-50 MLLMs require significantly fewer instances than Top-50 MLLMs. 
This implies that accurately ranking higher-performing MLLMs (Top-50) demands more instances,
while ranking lower-performing MLLMs (Bottom-50) can be achieved with fewer instances. Consequently, the redundancy of benchmark instances correlates strongly with the capability of the MLLMs being evaluated: \textbf{the stronger the MLLMs, the lower the redundancy of the benchmark instances}.

From the benchmark-specific results (\cref{fig:instances_bench}), 
the redundancy gap between Top-50 and Bottom-50 MLLMs remains consistent across different benchmarks. 
Further examination reveals considerable variation in redundancy levels between benchmarks. For example, in the Top-50 redundancy analysis, 
RealWorldQA~\cite{RealWorldQA} demonstrates relatively low redundancy, 
requiring nearly 80\% of the instances to reach saturation, 
while other benchmarks require far fewer. 
However, for Bottom-50 MLLMs, redundancy levels across benchmarks increase significantly, and the differences between them narrow. 
This illustrates that benchmark redundancy is more prominent when evaluating less capable MLLMs.

It is important to note that the conclusions above are based on the statistical analysis of mainstream benchmarks. 
Specialized benchmarks, with unique design goals or tasks, require case-by-case analyses to assess their instance redundancy accurately. 
Therefore, while these results provide general insights into redundancy trends for standard benchmarks, further evaluation is necessary for niche or task-specific benchmarks.

\begin{figure*}
    \centering
    \includegraphics[width=\linewidth]{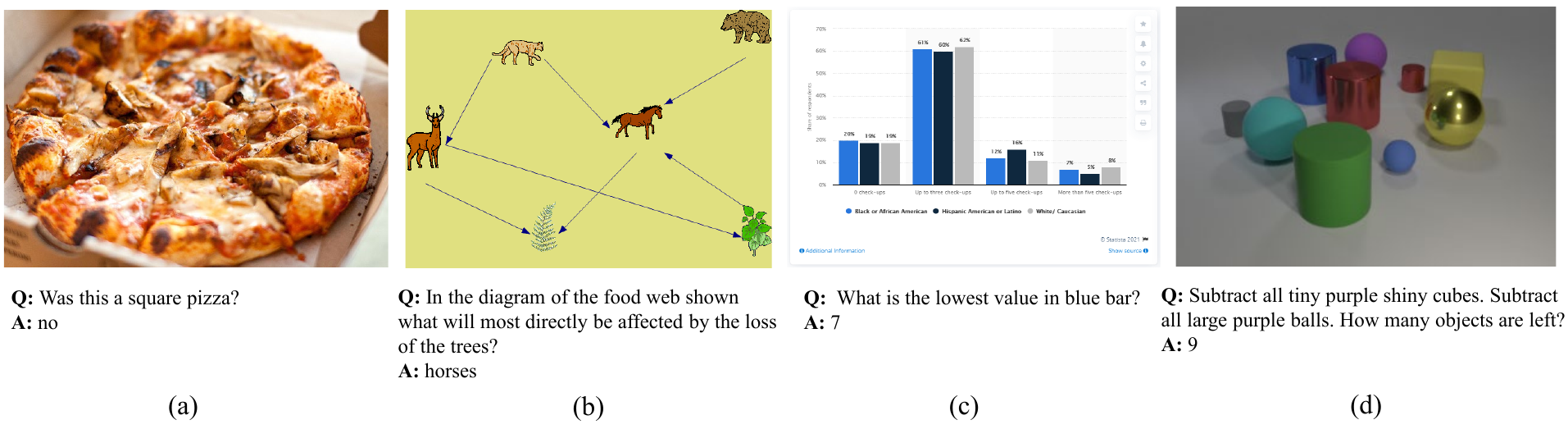}
    \caption{Examples of tasks excluded from the MathVista benchmark. (a), (b), and (c) showcase tasks derived from the \textit{general-vqa} category, including \textit{Scientific Figure Understanding}, \textit{General VQA}, and \textit{Chart/Table/Diagram QA}. Panel (d) presents questions extracted from the CLEVR dataset but categorized as \textit{math-targeted-vqa}.}
    \label{fig:append_mathvista}
    % \vspace{-10pt}
\end{figure*}

\subsection{Exploring Cross-Benchmark Redundancy}
\begin{figure}[t]
        \centering
        \includegraphics[width=\linewidth]{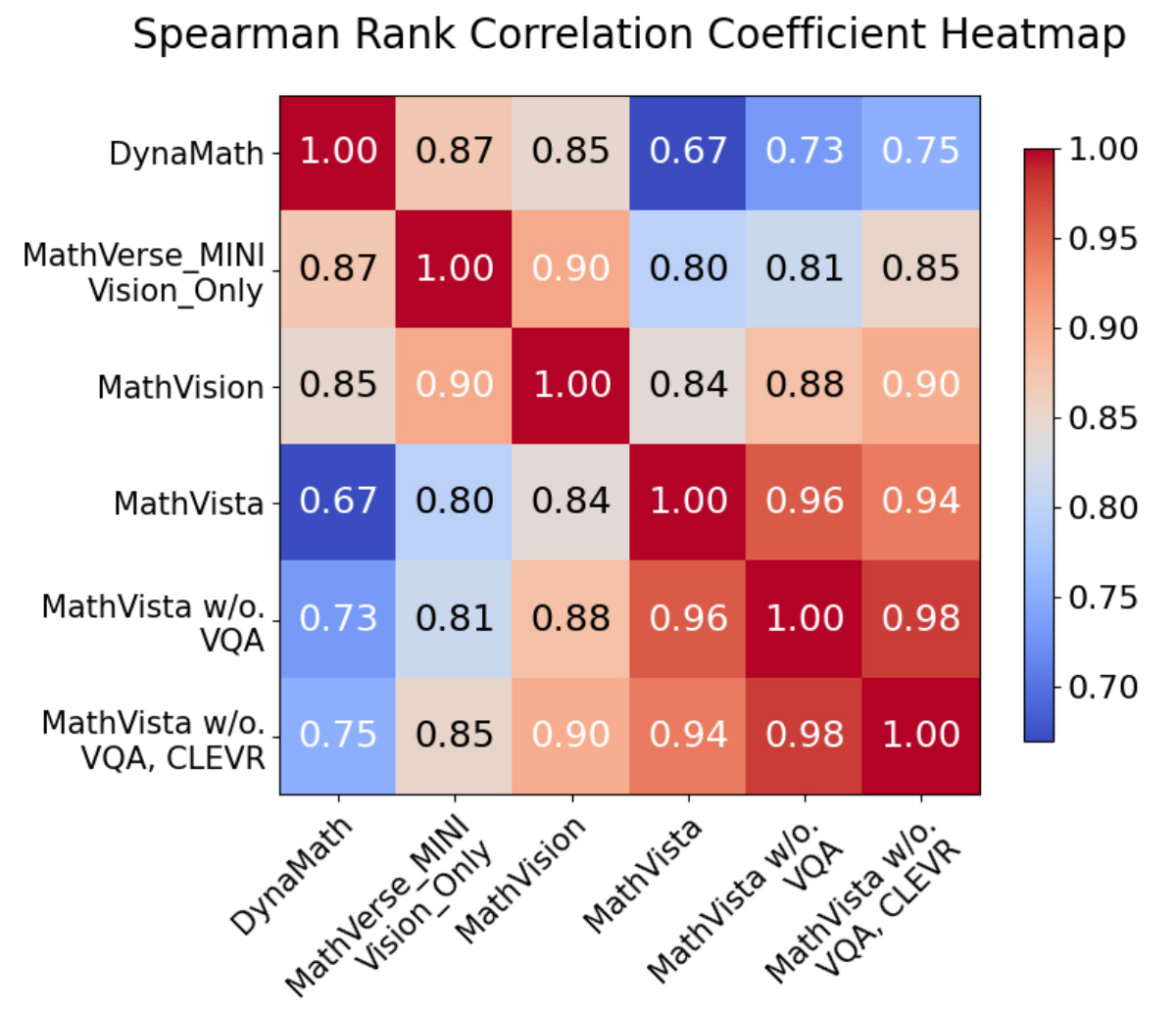}
        \caption{Cross-benchmark redundancy map. MathVision and MathVersion are more focused on the core domain of mathematics (with relatively higher redundancy across other math benchmarks), making them more suitable for benchmarking the mathematical capabilities of MLLMs in a narrow sense.}
        \label{fig:math_heatmap}
        \vspace{-10pt}
\end{figure}

To analyze cross-benchmark redundancy, we focus on the Math domain, 
specifically examining several popular mathematical benchmarks:
MathVista~\cite{lu2023mathvista}, MathVision~\cite{zhang2025mathverse}, MathVerse~\cite{wang2024measuring}, and DynaMath~\cite{zou2024dynamath}. 
We utilize the available evaluation results of 37 MLLMs listed on the OpenCompass Reasoning Leaderboard\footnote{\url{https://huggingface.co/spaces/opencompass/Open_LMM_Reasoning_Leaderboard}} and assess their ranking performance across these math benchmarks.
The corresponding heatmap is presented in \cref{fig:math_heatmap}.
The results reveal that, although all four benchmarks are designed to evaluate the mathematical abilities of MLLMs, 
the correlations between them are not particularly strong. 
Among them, MathVista~\cite{lu2023mathvista} exhibits the least redundancy, 
showing the lowest correlation with the other benchmarks. 
In contrast, MathVerse and MathVision demonstrate high redundancy, 
indicating a strong correlation with other benchmarks. 
These differences suggest varying levels of overlap in their evaluation focus areas.

To better understand the variability across benchmarks, we analyze their task distributions. 
While MathVerse and MathVision are focused on standard mathematical tasks, 
resulting in the highest correlation and substantial overlap with other benchmarks, 
MathVista includes 30\%-40\% of questions outside traditional mathematics, 
such as tasks related to \textit{Scientific Figure Understanding}, \textit{General VQA}, and \textit{Chart/Table/Diagram QA} (see \cref{fig:append_mathvista}(a)(b)(c) for examples). 
As discussed in \cref{sec2.3}, 
low redundancy can arise from unique elements specific to a domain or from irrelevant tasks, 
which we consider ``noise" within the dataset. 
For instance, general VQA tasks, while broadly useful, 
have limited relevance to assessing mathematical ability and contribute to this noise.
To quantify the impact, we remove general VQA tasks from MathVista and recalculate its redundancy with other benchmarks. 
After this refinement, the redundancy between MathVista and other mathematical benchmarks significantly increases, 
aligning more closely with their task profiles. 
Additionally, we identify and exclude \textit{CLEVR}-derived questions categorized as \textit{math-targeted vqa} within MathVista, 
which also have limited relevance to mathematical capabilities (examples in \cref{fig:append_mathvista}(d)).
This further increases overlap with specialized mathematical benchmarks, 
demonstrating that removing irrelevant tasks improves alignment and reduces noise.

Therefore, we propose the following principles for benchmark design within a domain:

\begin{itemize}[left=0pt, itemsep=0pt, topsep=0pt, parsep=0pt]
\item A benchmark intended to broadly assess model performance in one domain should demonstrate relatively high redundancy with other in-domain benchmarks, 
reflecting comprehensive coverage of diverse sub-capabilities.
\item A specialized benchmark should display lower redundancy with other benchmarks, 
focusing on distinct capabilities to fill the vacancy, complement broader assessments, and provide a unique perspective on specific topics in a domain.
\end{itemize}

\section{Conclusion}

In conclusion, this paper addresses the pervasive issue of redundancy in MLLM benchmarks, 
impacting both the effectiveness and efficiency of model evaluation. 
We identify redundancy at three levels: dimension, instance, and cross-benchmark redundancy,
and propose a framework with actionable guidelines to improve benchmark design. 
By promoting the independence of dimensions, optimizing instance counts, and ensuring purposeful redundancy within specific domains, 
our framework streamlines evaluations and enhances reliability. 
Case studies further demonstrate its utility in refining current practices, 
paving the way for more efficient and accurate MLLM assessments.

\newpage

\section{Limitations}
The limitations of this work is as follows:
\begin{itemize}[left=0pt, itemsep=0pt, topsep=0pt, parsep=0pt]
\item The assumption that MLLM performance rankings should show strong correlation when evaluating similar capabilities may not always hold. In some cases, performance on seemingly similar tasks could diverge due to subtle task differences, domain-specific nuances, or differences in model strengths. 
\item The use of correlation metrics (SRCC, PLCC, and R²) to quantify redundancy may be limited in capturing the full complexity of model performance across different tasks and domains. These metrics may not adequately account for differences in task difficulty, model behavior under various conditions, or the impact of outliers. 
\item The redundancy value is not fixed when using different selections of MLLMs for calculation. This bias could result in misleading conclusions about the redundancy or uniqueness of certain benchmarks.
\end{itemize}

\section{Acknowledgements}
This research was partly supported by grants of National Natural Science Foundation of China (NSFC, Grant No. 62171281), Science and Technology Commission of Shanghai Municipality (STCSM, Grant No. 20DZ1200203, 2021SHZDZX0102), National Key R\&D Program of China (No.2022ZD0161600), the Shanghai Postdoctoral Excellence Program (No.2023023), China Postdoctoral Science Fund (No.2024M751559), and Shanghai Artificial intelligence Laboratory.

\bibliography{main}

\appendix

\newpage

\section{Metrics Equation}
\label{sec:metrics}

To evaluate the consistency and accuracy of predictions, we employ three widely used metrics: the Spearman Rank Correlation Coefficient (SRCC), the Pearson Linear Correlation Coefficient (PLCC), and the Coefficient of Determination (\(R^2\)). These metrics provide complementary perspectives on model performance, capturing rank-based, linear, and variance-explained relationships, respectively. The mathematical definitions are detailed below.

1) The SRCC measures the rank-based relationship between predicted and true values. It is defined as:
\[
\text{SRCC} = 1 - \frac{6 \sum_{i=1}^n d_i^2}{n(n^2 - 1)},
\]
where:
\[
d_i = \text{rank}(x_i) - \text{rank}(y_i),
\]
and \(n\) is the number of data points. A higher SRCC indicates a stronger monotonic relationship between the rankings of predicted and ground truth values.

2) The PLCC quantifies the linear relationship between predicted and true values. It is computed as:
\[
\text{PLCC} = \frac{\sum_{i=1}^n (x_i - \bar{x})(y_i - \bar{y})}{\sqrt{\sum_{i=1}^n (x_i - \bar{x})^2} \sqrt{\sum_{i=1}^n (y_i - \bar{y})^2}},
\]
where:
\begin{itemize}[left=0pt, itemsep=0pt, topsep=0pt, parsep=0pt]
    \item \(x_i\) and \(y_i\) are the data points,
    \item \(\bar{x}\) and \(\bar{y}\) are the means of \(x\) and \(y\), respectively.
\end{itemize}
A higher PLCC indicates a stronger linear relationship between predicted and ground truth values.

3) The \(R^2\) score represents the proportion of variance in the ground truth values that is explained by the predictions. It is defined as:
\[
R^2 = 1 - \frac{\sum_{i=1}^n (y_i - \hat{y}_i)^2}{\sum_{i=1}^n (y_i - \bar{y})^2}
\]
where:
\begin{itemize}[left=0pt, itemsep=0pt, topsep=0pt, parsep=0pt]
    \item \(y_i\) are the ground truth values,
    \item \(\hat{y}_i\) are the predicted values,
    \item \(\bar{y}\) is the mean of the ground truth values.
\end{itemize}
An \(R^2\) score closer to 1 indicates a better fit between the predictions and the ground truth.

% These metrics collectively provide a comprehensive evaluation of the model's performance, addressing rank consistency, linear correlation, and variance explanation.

\section{Extra Dimensions Redundancy Maps}
\label{app:extra}
\begin{figure*}[tbp]
    \centering
    \begin{subfigure}[b]{.32\linewidth} 
        \centering
        \includegraphics[width=\linewidth]{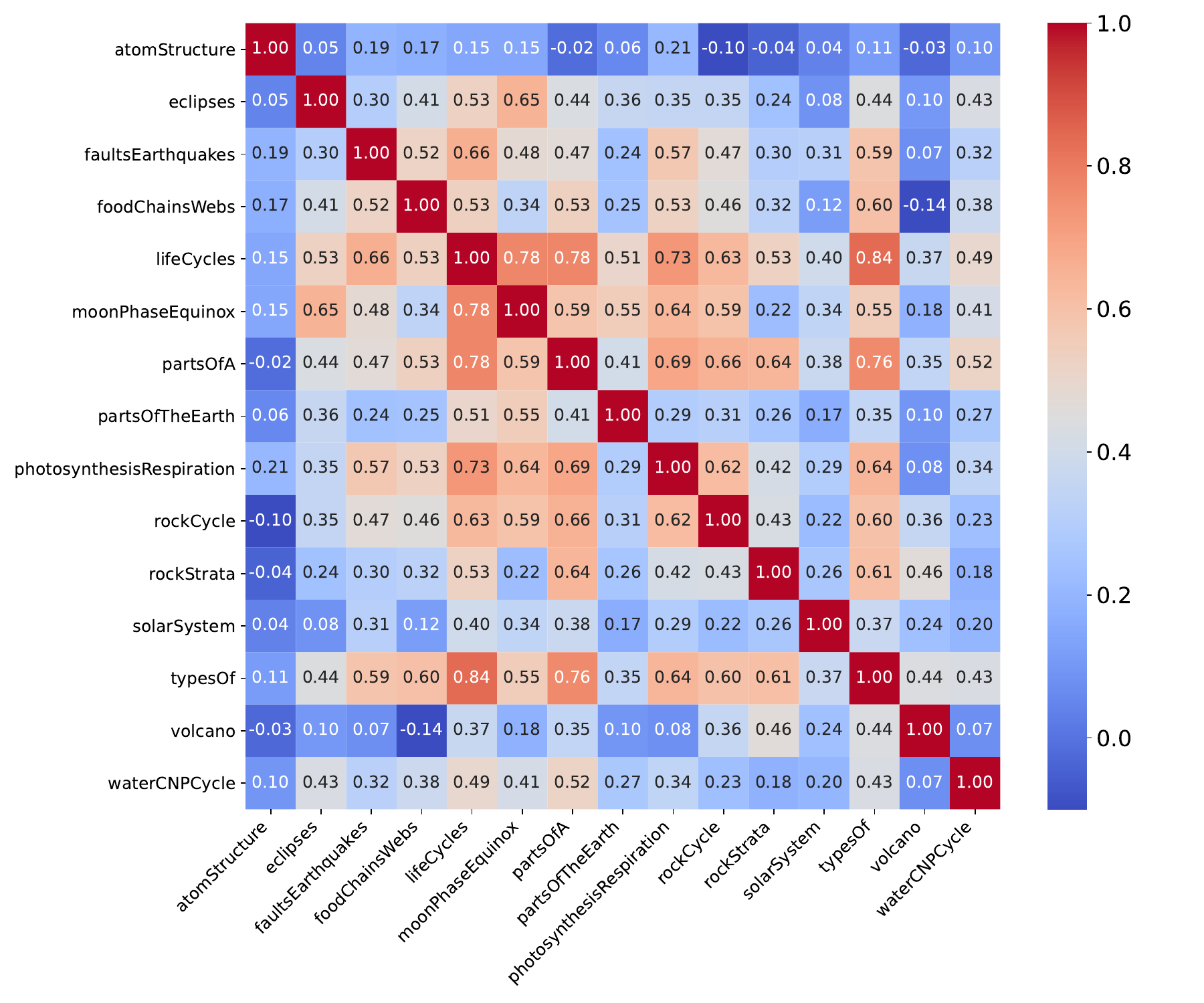}
        \caption{50$^+$ SRCC dimensions redundancy.}
        \label{fig:AI2D_SRCC_50_heat}
    \end{subfigure}
    \hfill
    \begin{subfigure}[b]{.32\linewidth} 
        \centering
        \includegraphics[width=\linewidth]{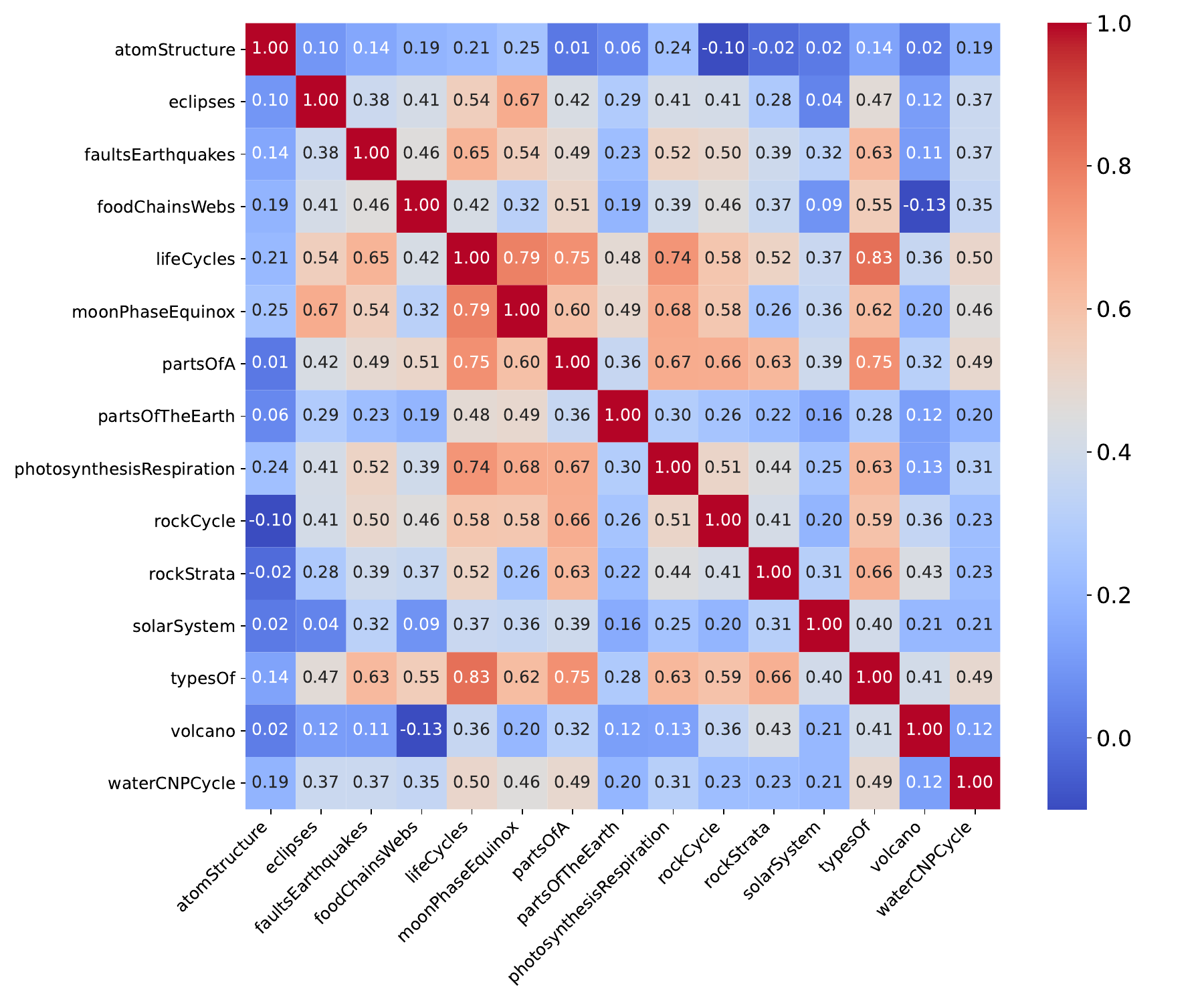}
        \caption{50$^+$ PLCC dimensions redundancy.}
        \label{fig:AI2D_PLCC_50_heat}
    \end{subfigure}
    \hfill
    \begin{subfigure}[b]{.32\linewidth} 
        \centering
        \includegraphics[width=\linewidth]{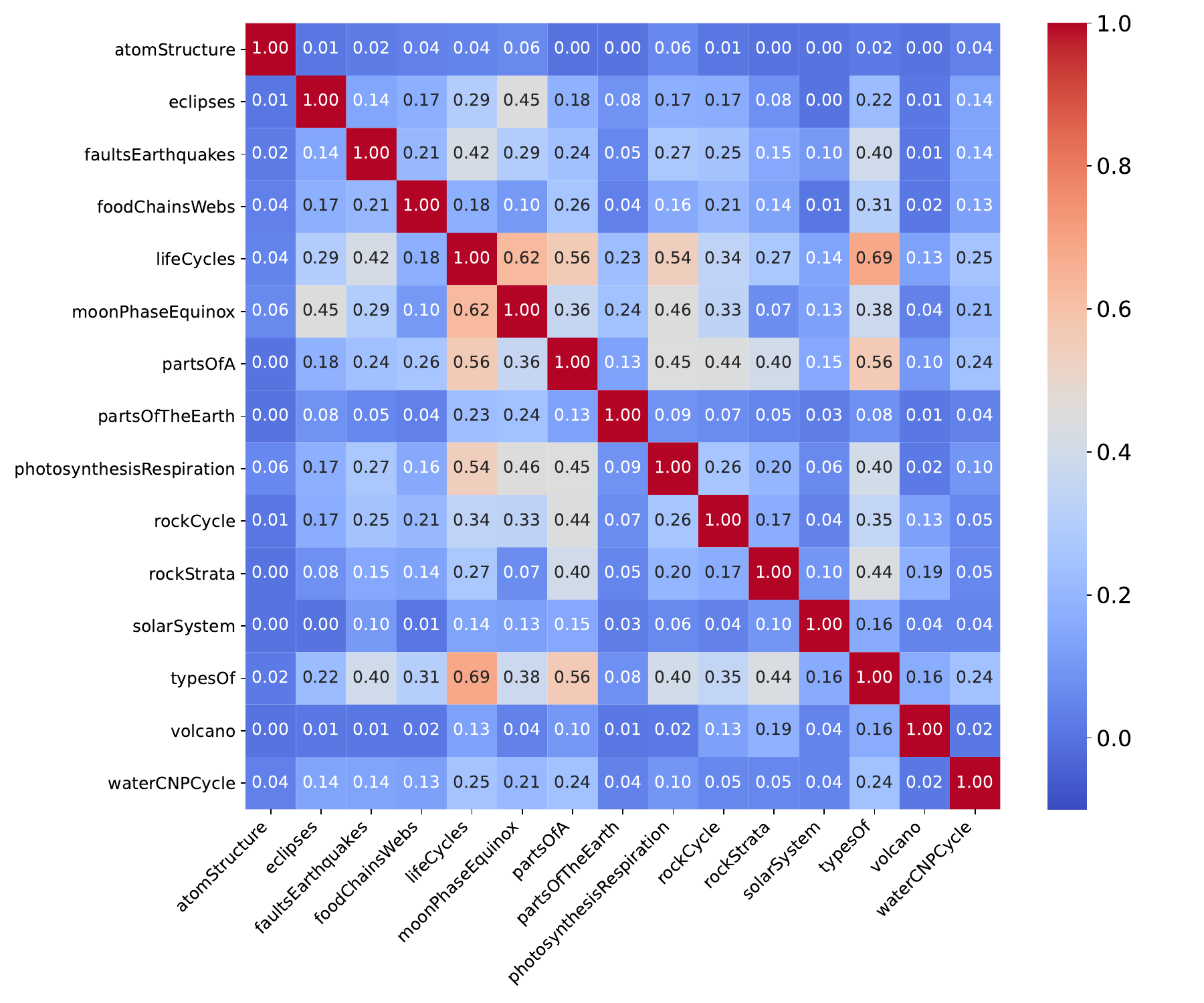}
        \caption{50$^+$ R2 dimensions redundancy.}
        \label{fig:AI2D_R2_50_heat}
    \end{subfigure}
    \begin{subfigure}[b]{.32\linewidth}  
        \centering
        \includegraphics[width=\linewidth]{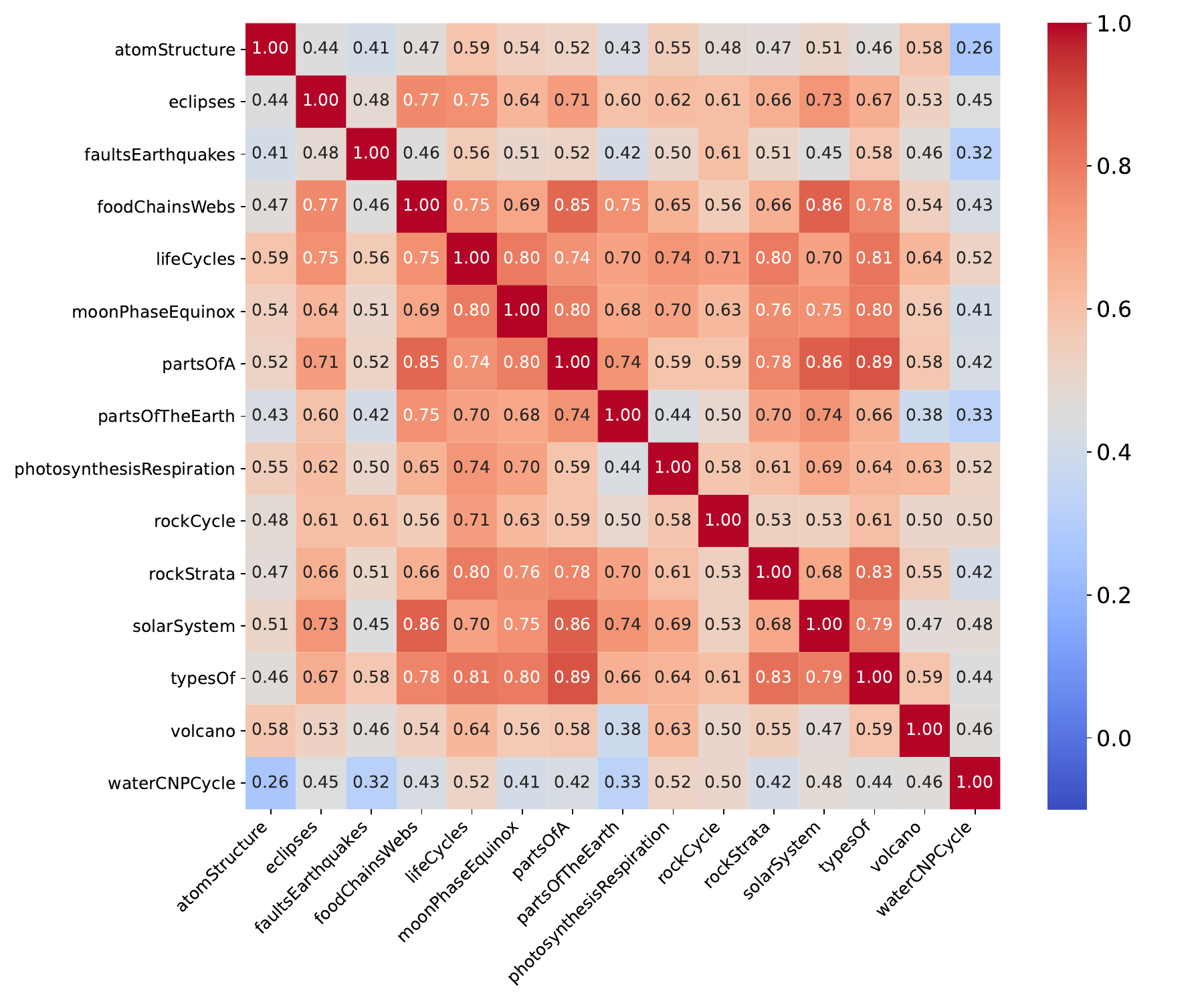}
        \caption{50$^-$ SRCC dimensions redundancy.}
        \label{fig:AI2D_SRCC_-50_heat}
    \end{subfigure}
    \hfill
    \begin{subfigure}[b]{.32\linewidth}  
        \centering
        \includegraphics[width=\linewidth]{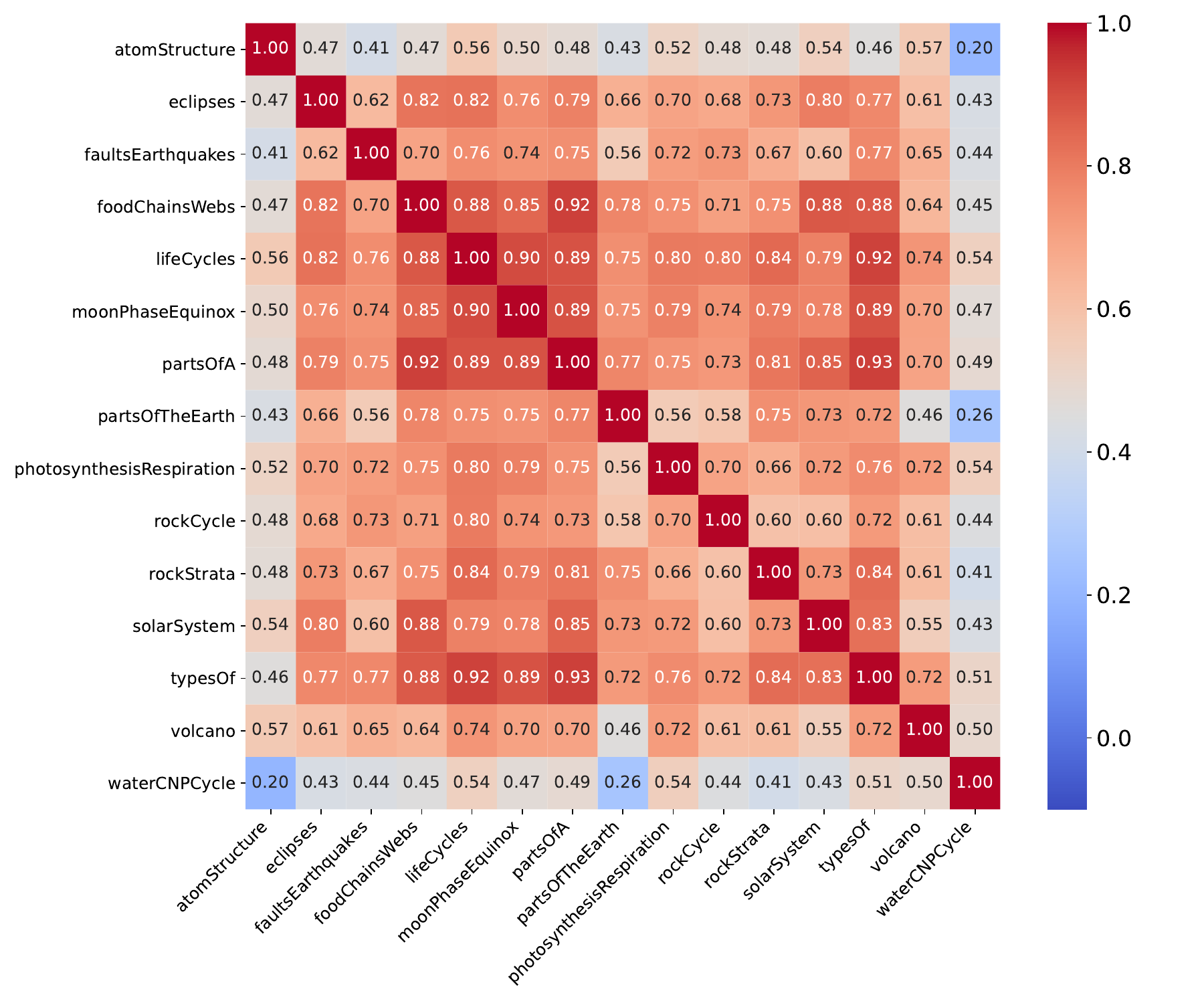}
        \caption{50$^-$ PLCC dimensions redundancy.}
        \label{fig:AI2D_PLCC_-50_heat}
    \end{subfigure}
    \hfill
    \begin{subfigure}[b]{.32\linewidth}  
        \centering
        \includegraphics[width=\linewidth]{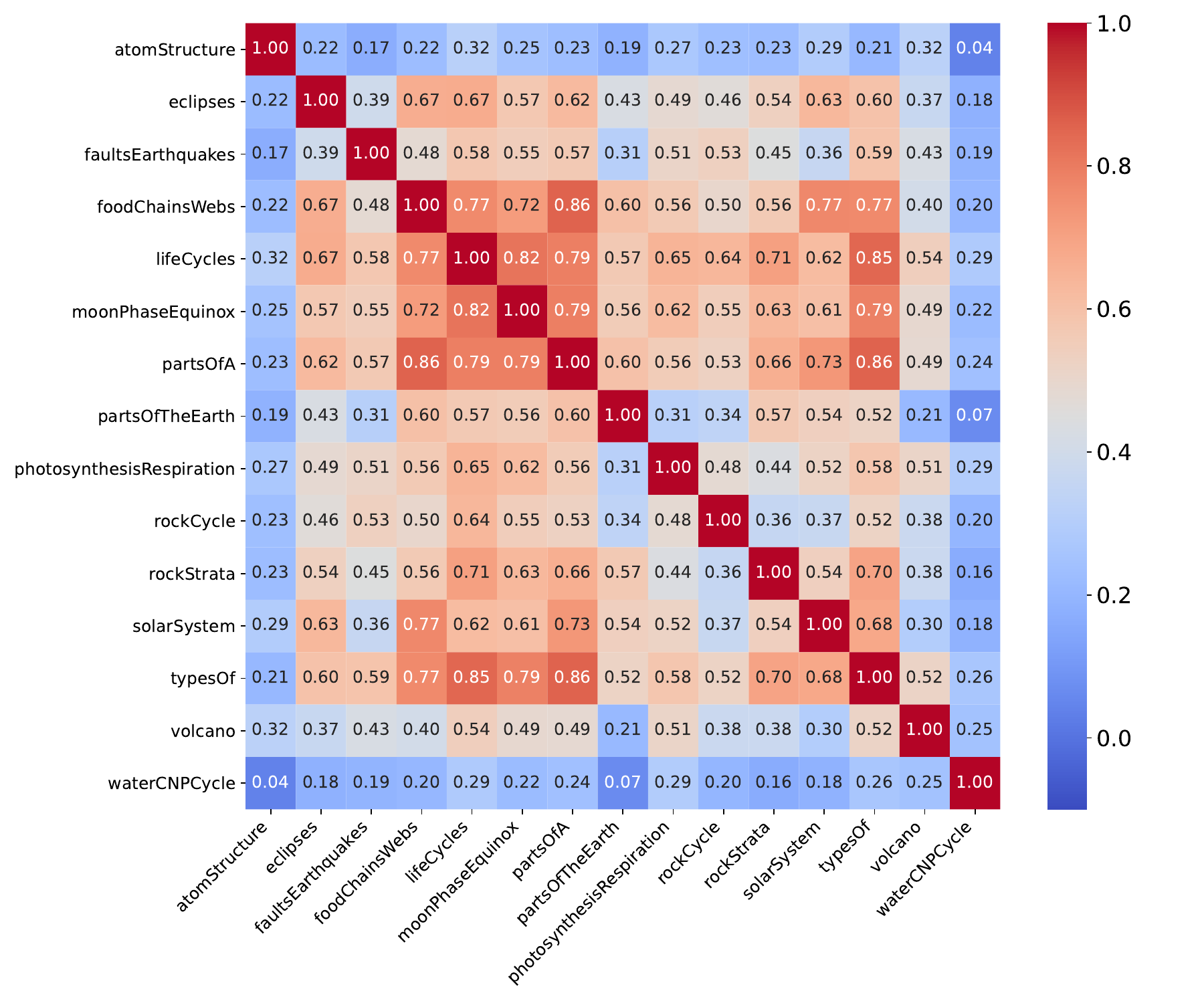}
        \caption{50$^-$ R2 dimensions redundancy.}
        \label{fig:AI2D_R2_-50_heat}
    \end{subfigure}
    \caption{Visualizations of dimensions redundancy for AI2D~\cite{kembhavi2016diagram} on Top-50 and Bottom-50 MLLMs (\textit{marked as 50$^+$ and 50$^-$}).}
    \label{fig:ai2d_heat}
\end{figure*}

\begin{figure*}[tbp]
    \centering
    \begin{subfigure}[b]{.32\linewidth} 
        \centering
        \includegraphics[width=\linewidth]{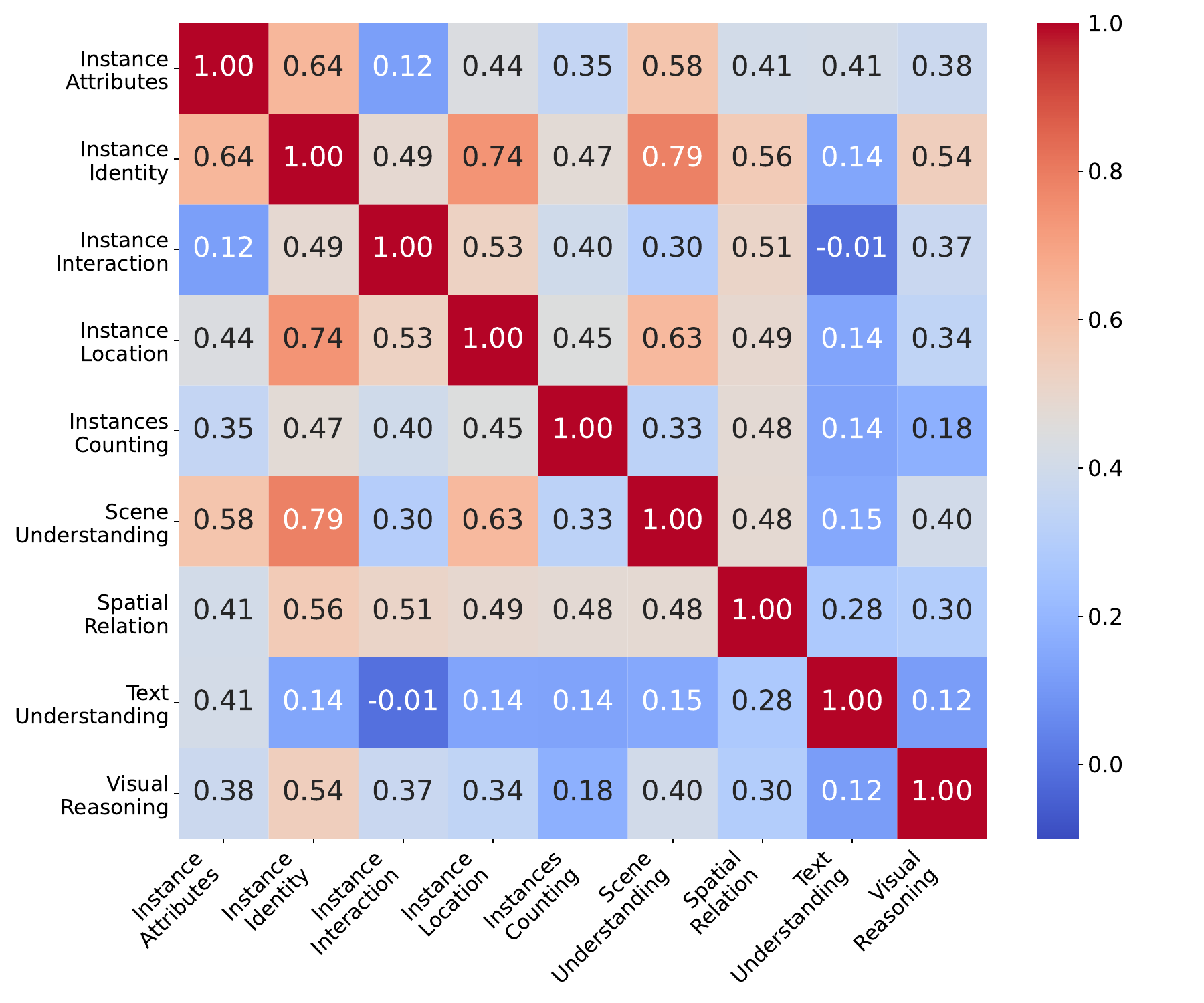}
        \caption{50$^+$ SRCC dimensions redundancy.}
        \label{fig:SEEDBench_SRCC_50_heat}
    \end{subfigure}
    \hfill
    \begin{subfigure}[b]{.32\linewidth} 
        \centering
        \includegraphics[width=\linewidth]{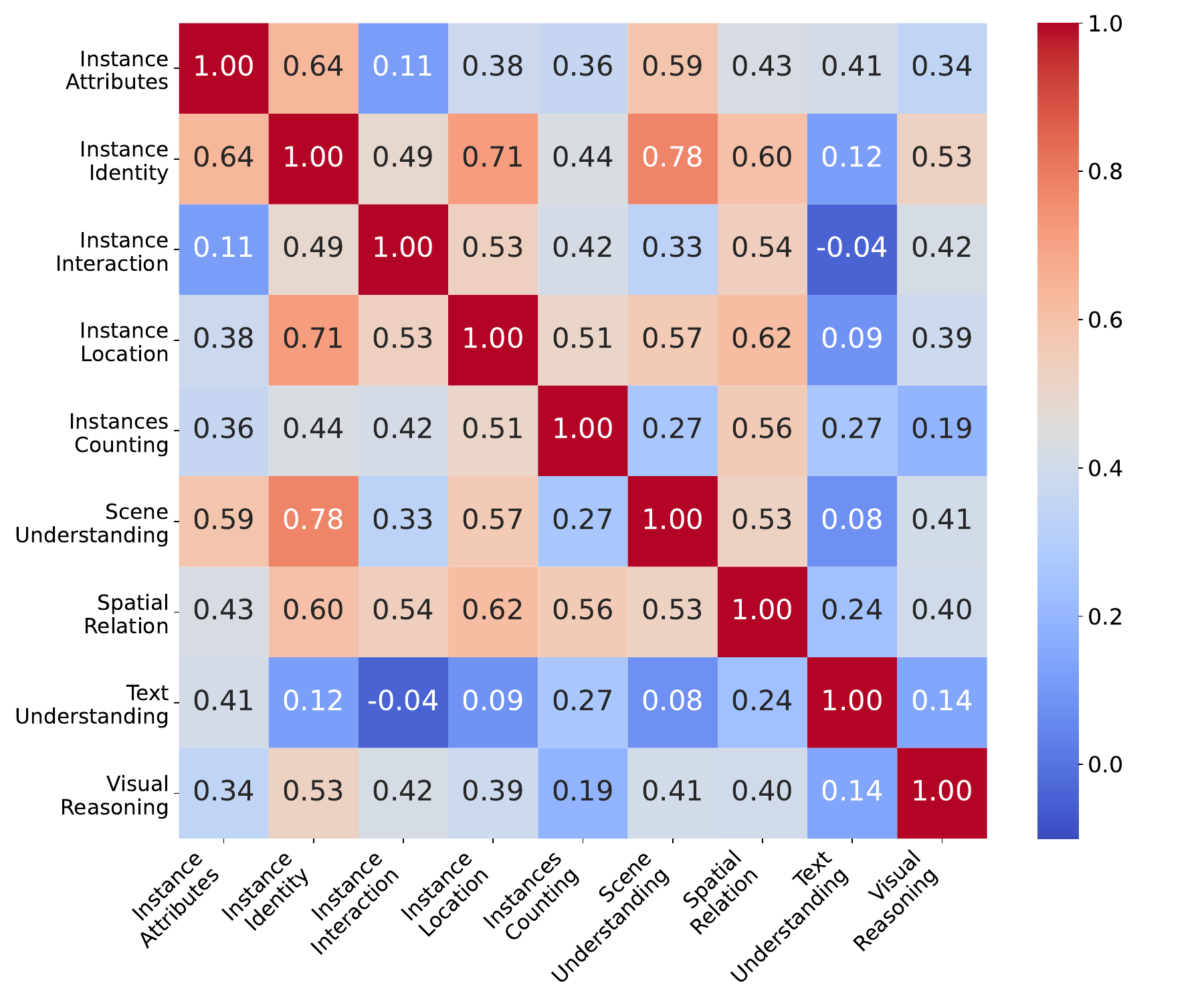}
        \caption{50$^+$ PLCC dimensions redundancy.}
        \label{fig:SEEDBench_PLCC_50_heat}
    \end{subfigure}
    \hfill
    \begin{subfigure}[b]{.32\linewidth} 
        \centering
        \includegraphics[width=\linewidth]{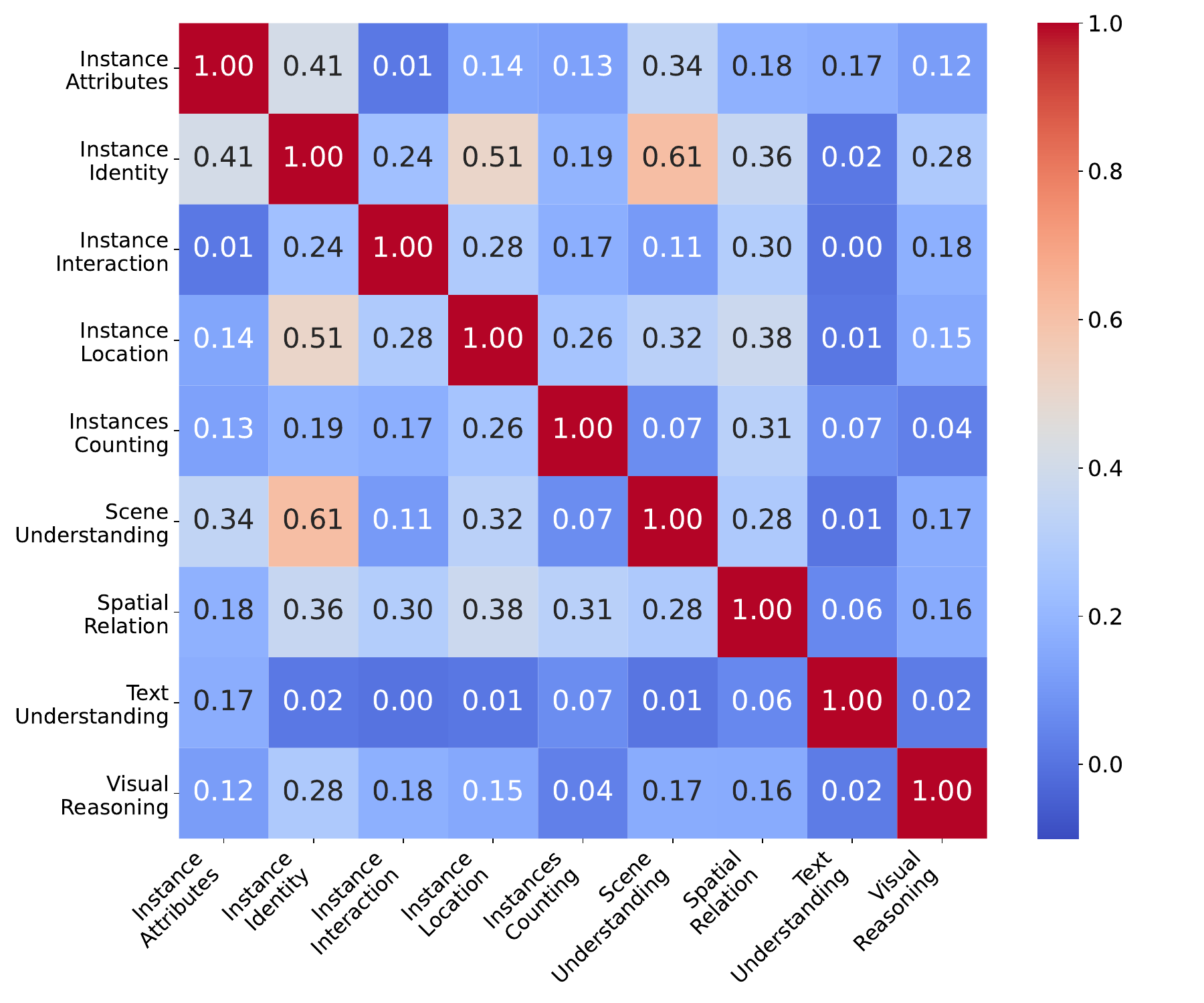}
        \caption{50$^+$ R2 dimensions redundancy.}
        \label{fig:SEEDBench_R2_50_heat}
    \end{subfigure}
    \begin{subfigure}[b]{.32\linewidth}  
        \centering
        \includegraphics[width=\linewidth]{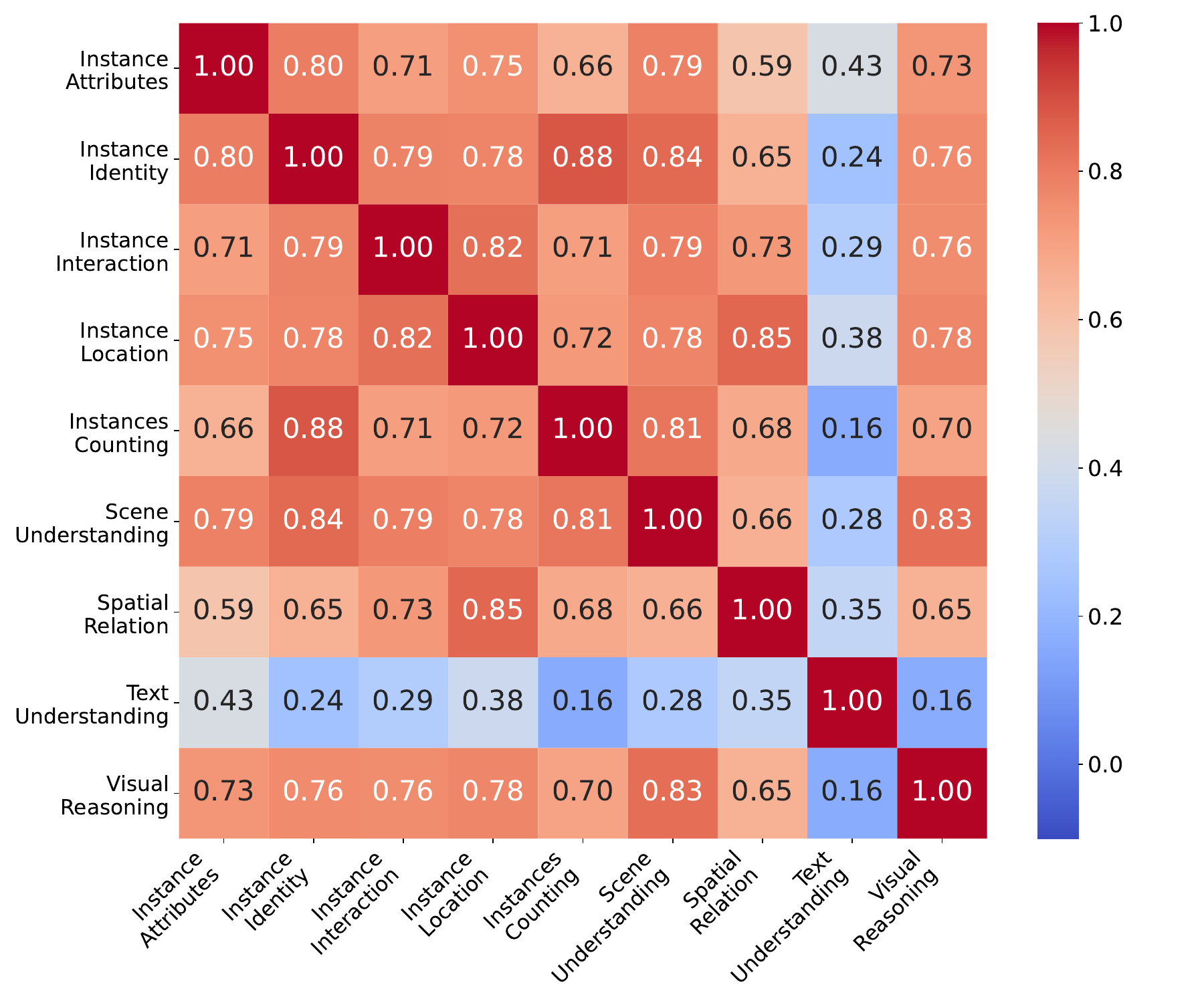}
        \caption{50$^-$ SRCC dimensions redundancy.}
        \label{fig:SEEDBench_SRCC_-50_heat}
    \end{subfigure}
    \hfill
    \begin{subfigure}[b]{.32\linewidth}  
        \centering
        \includegraphics[width=\linewidth]{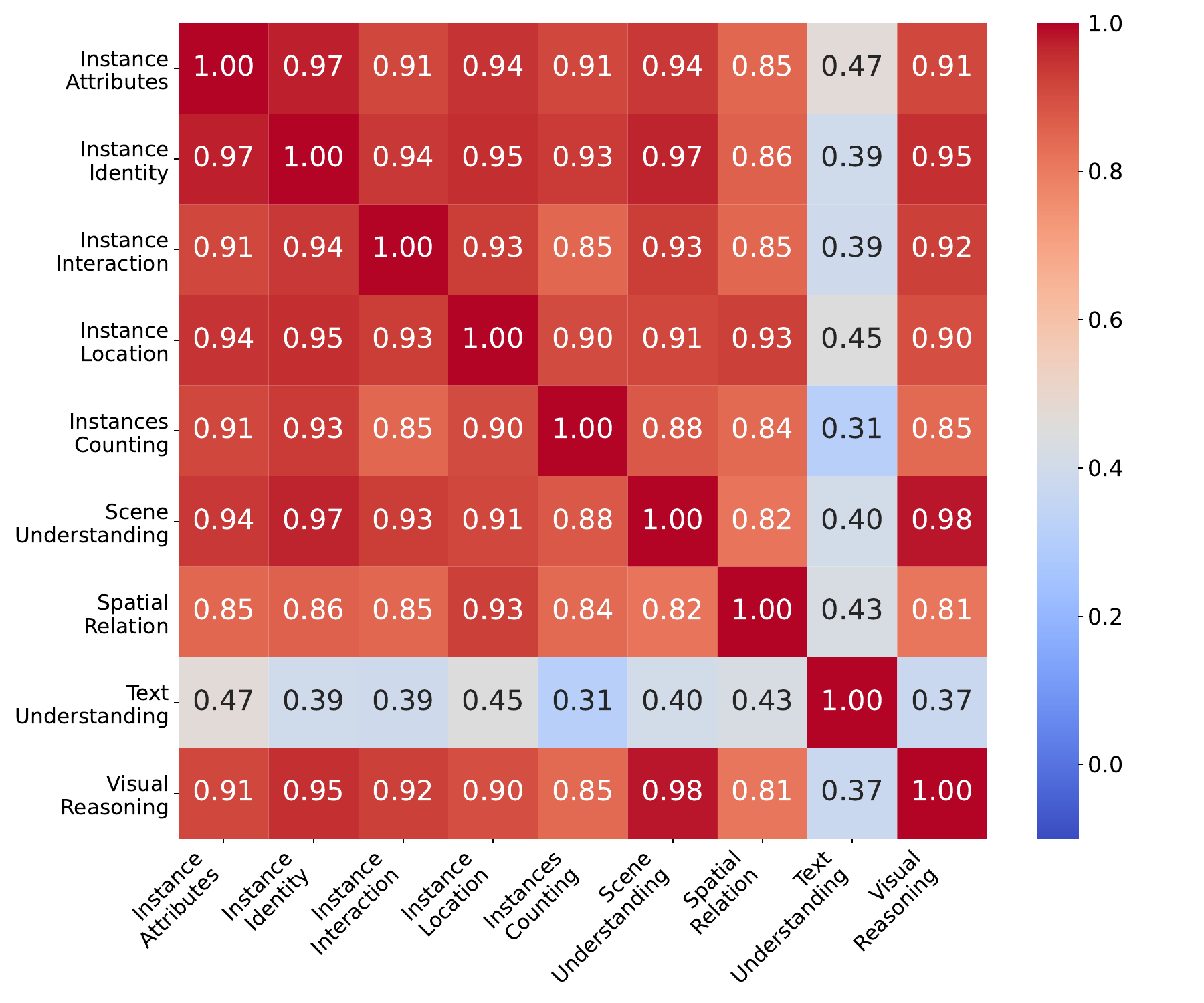}
        \caption{50$^-$ PLCC dimensions redundancy.}
        \label{fig:SEEDBench_PLCC_-50_heat}
    \end{subfigure}
    \hfill
    \begin{subfigure}[b]{.32\linewidth}  
        \centering
        \includegraphics[width=\linewidth]{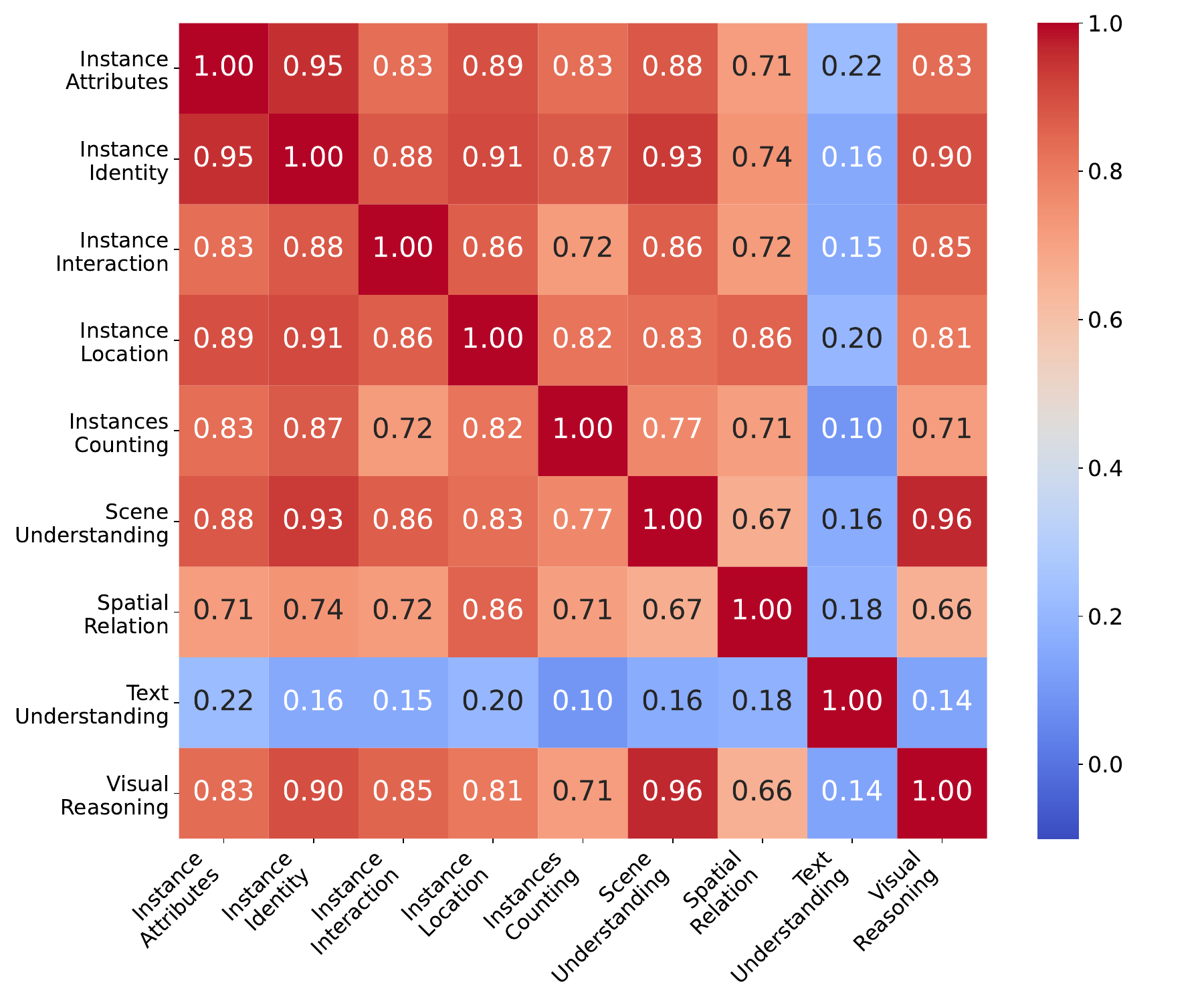}
        \caption{50$^-$ R2 dimensions redundancy.}
        \label{fig:SEEDBench_R2_-50_heat}
    \end{subfigure}
    \caption{Visualizations of dimensions redundancy for SEED-Bench~\cite{li2024seed} on Top-50 and Bottom-50 MLLMs (\textit{marked as 50$^+$ and 50$^-$}).}
    \label{fig:seedbench_heat}
\end{figure*}

We present the dimension redundancy maps for AI2D~\cite{kembhavi2016diagram} and SEED-Bench~\cite{li2024seed}, as shown in Fig.~\ref{fig:ai2d_heat} and Fig.~\ref{fig:seedbench_heat}. 

\begin{enumerate}
    \item \textbf{Key Observations from the Redundancy Maps:}
    \begin{itemize}[left=0pt, itemsep=0pt, topsep=0pt, parsep=0pt]
        \item In Fig.~\ref{fig:ai2d_heat}, it is evident that the dimension `lifeCycles' exhibits the highest redundancy, particularly with `typesOf'.
        \item Similarly, in Fig.~\ref{fig:seedbench_heat}, the `Instance Identity' dimension shows the highest redundancy and is most closely related to `Scene Understanding'.
    \end{itemize}

    \item \textbf{Trends in Top-50 vs. Bottom-50 Redundancy:}
    \begin{itemize}[left=0pt, itemsep=0pt, topsep=0pt, parsep=0pt]
        \item A clear pattern emerges when comparing the Top-50 and Bottom-50 redundancy maps. Nearly all Bottom-50 dimensions display significantly higher redundancy than their Top-50 counterparts. This observation supports our conclusion that \textbf{dimensions tend to exhibit greater redundancy for Bottom-50 MLLMs compared to Top-50 MLLMs}.
        \item This phenomenon can be attributed to the overall underperformance of Bottom-50 MLLMs across various capabilities. As these models begin to improve, enhancements in their foundational abilities often lead to simultaneous progress across multiple dimensions. This results in a high degree of similarity in performance rankings, contributing to elevated dimensional redundancy.
        \item In contrast, Top-50 MLLMs already possess relatively strong foundational capabilities. As a result, more challenging tasks across different dimensions introduce greater differentiation, reducing redundancy and creating more distinct performance profiles.
    \end{itemize}

    \item \textbf{Implications for Redundancy Analysis:}
    \begin{itemize}[left=0pt, itemsep=0pt, topsep=0pt, parsep=0pt]
        \item To ensure a reasonable and accurate evaluation during redundancy analysis, it is crucial to exclude MLLMs with consistently poor performance. Including such models could skew the analysis by disproportionately inflating redundancy, as their universal underperformance does not provide meaningful insights into inter-dimensional relationships.
    \end{itemize}
\end{enumerate}

\section{Redundancy Practice Recommendations}

To ensure benchmarks are reliable and efficient, 
we recommend incorporating redundancy detection into the benchmark design process after its initial testing on a set of MLLMs. 
This critical step identifies potential redundancies across dimensions/instances/cross-benchmark overlaps, leading to more precise and meaningful evaluations.

\begin{enumerate}
\item \textbf{Dimension Redundancy Check. }

Calculate the dimensional redundancy within the benchmark, 
with particular attention to dimensions exhibiting overall high redundancy. 
Analyze the redundancy heatmap to identify pairs of dimensions with exceptionally strong correlations, 
as these may indicate overlapping capabilities being assessed. 
For such cases, evaluate whether these dimensions are truly necessary or whether they assess similar or redundant skills.

\item
\textbf{Instance Redundancy Check. }

Compute the instance redundancy curve to determine whether a smaller subset of benchmark instances can produce results comparable to the full instance set. 
If significant instance redundancy is identified, 
the benchmark should be reviewed, and redundant instances should be reduced. 
This not only streamlines the evaluation process but also optimizes resource usage without compromising the accuracy of results.

\item
\textbf{Cross-benchmark Redundancy Check. }

If the benchmark is intended to serve as a representative for a specific domain, 
measure its cross-benchmark redundancy relative to other benchmarks within the domain. 
Higher redundancy indicates stronger representativeness, 
making it a reliable choice for tasks requiring domain coverage. 
Conversely, if the goal is to fill a vacancy in the specific domain (\textit{e.g.}, focusing on a specific topic in mathematics that is not covered by previous benchmarks)
maintaining low redundancy is a more favorable choice.
For use cases focusing on core capabilities within a specific domain under limited resources, 
it is recommended to select the benchmark with the highest cross-benchmark redundancy. 
This ensures that the benchmark comprehensively covers the essential skills while minimizing unnecessary overlaps.
\end{enumerate}

\end{document}